\documentclass{article}

\PassOptionsToPackage{numbers, compress}{natbib}

\usepackage[preprint]{neurips_2025}

\usepackage[utf8]{inputenc} 
\usepackage[T1]{fontenc}    
\usepackage{hyperref}       
\usepackage{url}            
\usepackage{booktabs}       
\usepackage{amsfonts}       
\usepackage{nicefrac}       
\usepackage{microtype}      
\usepackage{graphicx}

\usepackage[numbers]{natbib}
\usepackage{times}
\usepackage{latexsym}
\usepackage{subcaption}  
\usepackage[normalem]{ulem}
\usepackage{booktabs}
\usepackage{multirow}
\usepackage{makecell}
\useunder{\uline}{\ul}{}
\usepackage[table,xcdraw]{xcolor}
\usepackage{CJKutf8} 
\usepackage{colortbl}
\usepackage{color}
\usepackage[most]{tcolorbox}
\usepackage{arydshln}
\usepackage{wrapfig}
\usepackage{enumitem}
\usepackage{pifont}
\usepackage{rotating}

\newtcolorbox[list inside=prompt,auto counter,number within=section]{prompt}[1][]{
    colbacktitle=black!60,
    coltitle=white,
    fontupper=\footnotesize,
    boxsep=5pt,
    left=0pt,
    right=0pt,
    top=0pt,
    bottom=0pt,
    boxrule=1pt,
    #1,
}

\newcommand{\cmark}{\textcolor{green}{\ding{51}}}  
\newcommand{\xmark}{\textcolor{red}{\ding{55}}}    

\definecolor{ForestGreen}{rgb}{0.13, 0.55, 0.13} 
\definecolor{WildStrawberry}{rgb}{1.0, 0.26, 0.64}
\definecolor{AirForceblue}{rgb}{0.36, 0.54, 0.66}
\definecolor{aliceblue}{rgb}{0.94, 0.97, 1.0}
\newcommand{\increase}[1]{
	\fontsize{8pt}{0.5em}\selectfont\color{ForestGreen}{$\uparrow$~\textbf{#1}}
}
\newcommand{\decrease}[1]{
	\fontsize{8pt}{0.5em}\selectfont\color{WildStrawberry}{$\downarrow$~\textbf{#1}}
}

\title{AgriEval: A Comprehensive Chinese Agricultural Benchmark for Large Language Models}

\author{
	Lian Yan$^{1*}$ \quad Haotian Wang$^{1*}$ \quad Chen Tang$^{2}$\thanks{Equal contribution} \quad Haifeng Liu$^{1}$ \quad Tianyang Sun$^{1}$ \\ \textbf{Liangliang Liu}$^1$ \quad \textbf{Yi Guan}$^1$ \quad  \textbf{Jingchi Jiang}$^1$\thanks{Corresponding Author} \\
	$^{1}$ Harbin Institute of Technology, $^{2}$ MemTensor (Shanghai) Technology Co., Ltd.  \\
	\texttt{yanlian0216@163.com},  \texttt{jiangjingchi@hit.edu.cn} \\
}

\begin{document}

\maketitle

\begin{abstract}
In the agricultural domain, the deployment of large language models (LLMs) is hindered by the lack of training data and evaluation benchmarks. To mitigate this issue, we propose AgriEval, the first comprehensive Chinese agricultural benchmark with three main characteristics: (1) \textit{Comprehensive Capability Evaluation.} AgriEval covers six major agriculture categories and 29 subcategories within agriculture, addressing four core cognitive scenarios—memorization, understanding, inference, and generation. (2) \textit{High-Quality Data.} The dataset is curated from university-level examinations and assignments, providing a natural and robust benchmark for assessing the capacity of LLMs to apply knowledge and make expert-like decisions. (3) \textit{Diverse Formats and Extensive Scale.} AgriEval comprises 14,697 multiple-choice questions and 2,167 open-ended question-and-answer questions, establishing it as the most extensive agricultural benchmark available to date. We also present comprehensive experimental results over 51 open-source and commercial LLMs. The experimental results reveal that most existing LLMs struggle to achieve 60\% accuracy, underscoring the developmental potential in agricultural LLMs. Additionally, we conduct extensive experiments to investigate factors influencing model performance and propose strategies for enhancement. AgriEval is available at \url{https://github.com/YanPioneer/AgriEval/}.
\end{abstract}

\section{Introduction}\label{introduction}
The rapid development of large language models (LLMs) has enabled new applications in smart agriculture \cite{tzachor2023large, LI2024109032, SHAHRIAR2025101787, KUSKA2024108924}, such as knowledge-based Q\&A \cite{silva2023gpt}, cultivation planning \cite{peng2023agri}, and plant science \cite{agathokleous2024one,macnish2025application,yang2024pllama}. However, agriculture is a highly specialized domain with fragmented knowledge, diverse subfields, and decisions requiring biological and environmental reasoning. Open-domain LLMs, lacking sufficient agricultural pre-training and domain grounding, often produce factually incorrect or misleading outputs in this context.

To address these challenges, a dedicated benchmark is essential for systematically evaluating LLMs' capabilities in the agricultural domain. The proposal of such a benchmark not only reflects the performance and limitations of current models in agriculture but also provides valuable insights for the potential development and enhancement of training agriculture-specific LLMs. Existing benchmarks \cite{huang2024c,li-etal-2024-cmmlu,zhang2024cmmmu,hendrycks2020measuring} predominantly focus on general or semi-professional knowledge, with limited coverage of agricultural topics. These studies, when considered as benchmarks for Agricultural AI, have two significant limitations: (1) \textit{Extremely limited agriculture-related content} (typically <1.5\% of total questions); and (2) \textit{A lack of expert-level questions}, with most items focusing on basic knowledge (e.g., crop identification) rather than complex reasoning required for tasks such as precision disease diagnosis or pesticide formulation \cite{jiang2025knowledge}. This dual deficiency, both in knowledge breadth and professional depth, renders current benchmarks inadequate for assessing LLMs' true competency in agricultural applications, where domain-specific knowledge and precise reasoning are critical for avoiding potentially serious real-world consequences.

In addition, benchmarks for Agricultural AI should introduce and account for more domain-specific challenges that extend beyond open-domain studies. For instance, regional diversity within the agricultural domain adds complexity that tests the generalization capabilities of LLMs. In particular, Chinese agriculture poses unique difficulties due to its regional heterogeneity, ecological diversity, and cultural specificity. Tasks such as pest control, crop breeding, and soil management are highly localized, while specialized subfields, such as traditional herbology and tea science, further broaden the domain scope. These factors should be incorporated into an Agricultural AI benchmark to comprehensively capture the breadth of agricultural knowledge and facilitate the fine-grained cognitive evaluation of LLMs.

\begin{figure*}[]
    \centering
   \makebox[\linewidth]{\includegraphics[width=\linewidth]{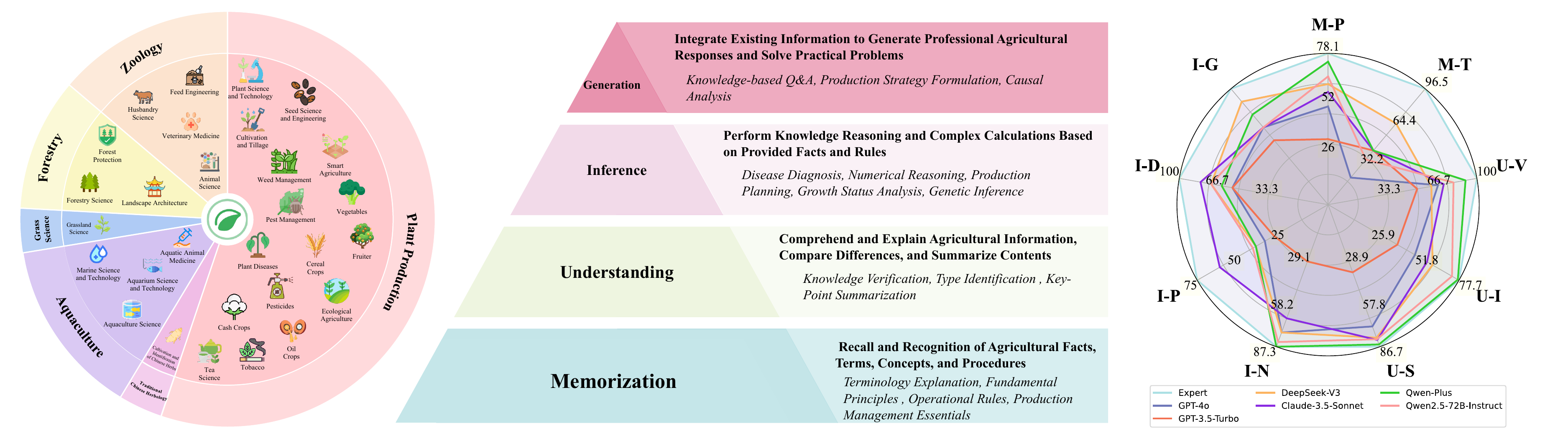}}
    \vspace{-5mm}
    \caption{\textit{Left}: Domains classification in AgriEval. \textit{Middle}: Cognitive ability classification in AgriEval. \textit{Right}: A brief overview of human and LLMs' performance on AgriEval.}
    \label{fig:intro}
\end{figure*}

To bridge the gap in LLM development and evaluation in the Agricultural domain, we propose \textbf{AgriEval}, the first large-scale benchmark for cognitive assessment in real-world Chinese agricultural scenarios.\footnote{The benchmark exclusively gathers Chinese agricultural data for two primary reasons: (1) The research group has strong connections with Chinese agricultural research teams, providing extensive data and access to professional experts and students. (2) The multilingual capabilities of current LLMs enable efficient and cost-effective translation, minimizing language barriers. An English-translated version of these benchmarks has also been released.} Developed under expert guidance, AgriEval covers six major categories and 29 subcategories (Figure~\ref{fig:intro} \textit{Left}). To meet the high specialization demands of agricultural production, we collect 14,697 multiple-choice questions (including single-answer, multiple-answer, true/false) and 2,167 Q\&A questions from college-level and professional exams. Inspired by Bloom's taxonomy \cite{seaman2011bloom,li2024lexeval} and real-world agricultural practices, AgriEval adopts a four-level cognitive framework—\textit{Memorization}, \textit{Understanding}, \textit{Inference}, and \textit{Generation}—further extended into 15 task-specific dimensions. This structure enables fine-grained evaluation of both knowledge breadth and reasoning depth in agricultural LLMs.

We conduct a comprehensive evaluation of 51 competitive LLMs on AgriEval, including nine commercial models and 42 open-source models spanning a wide range of architectures and parameter sizes. To assess their adaptability, we adopt multiple evaluation settings: zero-shot \cite{romera2015embarrassingly}, few-shot \cite{snell2017prototypical}, and chain-of-thought (CoT) \cite{wei2022chain,chu-etal-2024-navigate}. Furthermore, we explore option-shuffling, knowledge augmentation via retrieval-augmented generation (RAG) \cite{lewis2020retrieval}, and analyze across cognitive levels and task types to probe models' internal reasoning patterns and external knowledge dependency.

Our experiments demonstrate that current LLMs struggle to reach the performance of a human primary expert, with even the most capable model, Qwen-Plus, achieving only 63.21\% accuracy on AgriEval. Several key findings emerge from extensive experiments: (1) \textbf{Cognitive difficulty:} Performance varies significantly across cognitive levels, with numerical reasoning posing the greatest challenges. 
(2) \textbf{Positional sensitivity:} Most LLMs exhibit strong biases toward earlier options, performing poorly when correct answers appear in later positions.
(3) \textbf{Scaling effects:} Models under 7B parameters average 34.15\% accuracy; larger models generally perform better, though the trend is not strictly monotonic. 
(4) \textbf{Prompting strategies:} CoT boosts reasoning, particularly for complex tasks, while few-shot learning shows inconsistent gains. (5) \textbf{External knowledge integration:} RAG helps mitigate factual gaps in open-domain LLMs and improves performance on specialized tasks.

\section{Related Work}

Benchmarks play a critical role in evaluating model capabilities, akin to human-level examinations. Early benchmarks focused on task-specific objectives, such as machine translation \cite{bojar2014findings} and reading comprehension \cite{rajpurkar2018know}. With the emergence of LLMs, recent efforts have shifted toward evaluating general reasoning and embedded world knowledge \cite{zhang2024cmmmu, li2024lexeval, huang2024c, wang2024mmlu}. MMLU \cite{hendrycks2020measuring} consists of 15,908 multiple-choice questions across 57 subjects, spanning STEM and humanities, with varying levels of difficulty. Following this, multilingual and multimodal benchmarks have gained traction. For instance, C-Eval \cite{huang2024c} covers 52 Chinese-language disciplines, while CMMLU \cite{li-etal-2024-cmmlu} extends this to 67 subjects. CMMMU \cite{zhang2024cmmmu} introduces a multimodal benchmark across six core areas, including art, business, and science. MM-Vet \cite{yu2023mm} further examines multimodal understanding through tasks involving OCR, spatial reasoning, and mathematical problem-solving.

Despite their breadth, these benchmarks largely assess non-specialized or semi-specialized knowledge. As pointed out by LexEval \cite{li2024lexeval}, they offer limited insight into domains demanding expert-level understanding, such as medicine, law, finance, and agriculture. To address this gap, several domain-specific benchmarks have emerged: CMD \cite{wang2023cmb} for medicine, LexEval \cite{li2024lexeval} for legal reasoning, and Golden Touchstone \cite{lee2023rlaif} for financial analysis.

However, to our knowledge, no existing benchmark systematically evaluates LLMs in the agricultural domain. Given the domain's inherent complexity—spanning biological, ecological, and operational knowledge—and the added challenge of regional and cultural specificity in Chinese agriculture, a dedicated benchmark is urgently needed. Such a benchmark should capture the full spectrum of agricultural knowledge and support cognitive-level evaluation aligned with real-world applications.


\section{AgriEval}\label{s3}

\subsection{Task Overview}

\textbf{Motivation and Design Principles.}  
Unlike previous benchmarks for LLMs, the benchmarks for Agricultural AI should incorporate and address more domain-specific challenges beyond open-domain studies, and they provide three primary resources: (1) A taxonomy of tasks and corresponding datasets that represent the capabilities an LLM should possess to function as an agricultural expert; (2) A systematic evaluation method to assess various types of LLMs regarding these capabilities; (3) Comprehensive experiments that demonstrate how mainstream LLMs perform on the agricultural benchmark, offering insights for developing an agriculture-specific LLM. In essence, AgriEval aims to provide a hierarchical cognitive taxonomy of agricultural tasks aligned with real-world decision-making practices. This design facilitates both model diagnostics and targeted improvements for practical deployment. More details can be found in the Appendix~\ref{app_task_overview}.

\textbf{Domain Coverage.}  
Following the human-expert benchmark paradigm (e.g., C-Eval \cite{huang2024c}), AgriEval is curated from real examination questions designed for undergraduate and postgraduate students. With guidance from agricultural experts holding Ph.D. degrees in Agriculture within China's educational system, we align the domain taxonomy with China's official classification standards.\footnote{https://www.gov.cn/zhengce/zhengceku/2020-12/30/content\_5575377.htm} The benchmark spans six primary categories: \textit{\underline{P}lant \underline{P}roduction} (PP), \textit{\underline{Fore}stry} (Fore), \textit{\underline{G}rass \underline{S}cience} (GS), \textit{\underline{Aqua}culture} (Aqua), \textit{\underline{A}nimal \underline{S}cience and \underline{T}echnology} (AST), and \textit{\underline{T}raditional \underline{C}hinese \underline{H}erbology} (TCH). These are further divided into 29 subfields, such as plant protection, smart agriculture, and tea science (see Figure~\ref{fig:intro} (\textit{Left})).

\textbf{Cognitive Taxonomy.}
To assess both the breadth of knowledge and depth of reasoning in agricultural contexts, AgriEval introduces a four-level cognitive taxonomy inspired by Bloom's framework \cite{seaman2011bloom} and adapted from LexEval \cite{li2024lexeval}. The taxonomy consists of: (1) \textit{Memorization}, which evaluates the recall of facts, terms, and procedures; (2) \textit{Understanding}, which focuses on the ability to interpret, compare, and explain agricultural knowledge; (3) \textit{Inference}, which assesses reasoning and problem-solving based on domain knowledge; and (4) \textit{Generation}, which requires synthesizing information to produce professional, task-oriented responses. This hierarchical structure reflects the cognitive demands of real-world agricultural decision-making and supports fine-grained evaluation of LLMs. Complex tasks often span multiple levels, combining factual knowledge, reasoning, and domain-specific synthesis.

\begin{figure}[t]
  \centering
  \begin{minipage}[t]{0.66\textwidth}
    \vspace{0pt}
    \captionof{table}{AgriEval cognitive ability data statistics.}
    \label{tab:cognitive_data_statistics}
    \resizebox{\linewidth}{!}{
    \begin{tabular}{cccc}
        \toprule
        \textbf{Level}                 & \textbf{Task}                             & \textbf{\# Samples} & \textbf{Avg. Tokens} \\ \midrule
        \multirow{4}{*}{\underline{M}emorization}  & \underline{T}erminology Explanation (M-T)          &   125         & 101.85             \\
                                       & Fundamental \underline{P}rinciples (M-P)           &   6,077         & 82.51            \\
                                       & Operational \underline{R}ules (M-R)                &   116         & 89.7            \\
                                       & Production Management \underline{E}ssentials (M-E) &   880         & 98.21            \\ \midrule
        \multirow{3}{*}{\underline{U}nderstanding} & Knowledge \underline{V}erification (U-V)           &    1,961        & 45.75            \\
                                       & Type \underline{I}dentification (U-I)              &   2,253         &  80.3           \\
                                       & Key-Point \underline{S}ummarization (U-S)          &   1,324         &  103.32           \\ \midrule
        \multirow{5}{*}{\underline{I}nference}     & Production \underline{P}lanning (I-P)              & 471           & 95.34            \\ 
                                       & \underline{N}umerical Reasoning (I-N)              & 707           & 122.09            \\
                                       & Disease \underline{D}iagnosis (I-D)                & 403           & 114.7            \\
                                       & Growth \underline{S}tatus Analysis (I-S)           & 273           &  163.36           \\ 
                                       & \underline{G}enetic Inference (I-G)           & 107           &  125.37           \\ \midrule
        \multirow{3}{*}{\underline{G}eneration}    & Knowledge-based \underline{Q}\&\underline{A} (G-QA)            & 1,700           &  19.6           \\
                                       & \underline{P}roduction \underline{S}trategy Formulation (G-PS) & 325           &  41.72           \\
                                       & \underline{C}ausal \underline{A}nalysis (G-CA)                 & 142           & 22.92            \\ \bottomrule
        \end{tabular}
        }
    
  \end{minipage}
  \hfill
  \begin{minipage}[t]{0.32\textwidth}
  \vspace{3pt}
    \centering
    \includegraphics[width=\linewidth]{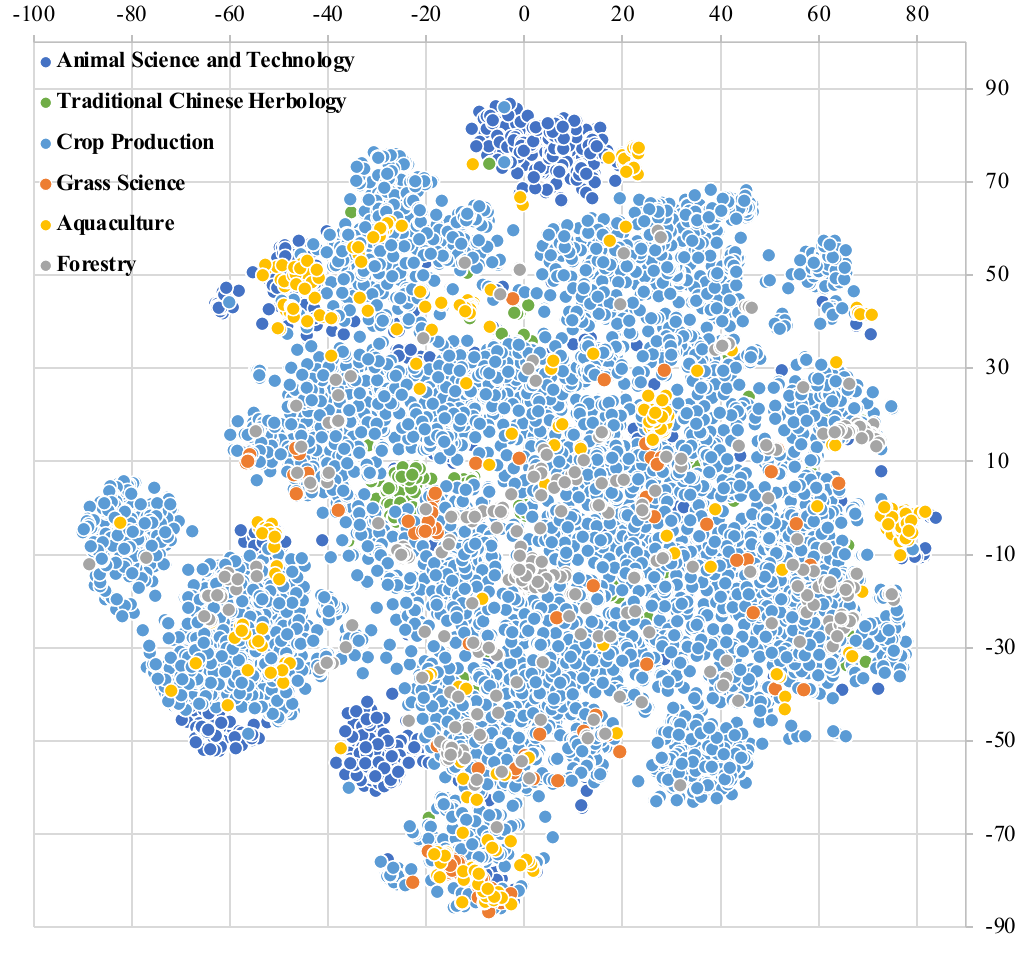}
    \captionof{figure}{Question representation via BERT encoding and t-SNE dimensionality reduction.}
    \label{fig:semantic_diversity}
  \end{minipage}%
\end{figure}

\subsection{Data Collection}\label{s3-data-collection}
We recruit two agricultural experts from partner agricultural university laboratories, both with advanced educational backgrounds, to collect domain-relevant examination materials from undergraduate and graduate-level assessments. Data sources include publicly available mock exams, graduate admission websites, and past exam materials publicly shared by students at top Chinese universities. All materials are originally in Word or PDF format. We collect over 500 documents and manually filter them based on difficulty, domain relevance, and alignment with real-world agricultural tasks, ultimately retaining 400 documents. The entire process spans approximately 1.0 months, with annotators compensated at 50 CNY per hour. Additional details on data sources and licensing are provided in the Appendix~\ref{app_data_license}.

\subsection{Data Annotation and Verification}\label{s3-data-verification}

To standardize the collected materials, all examination materials undergo a systematic digitization and structuring process. Source files in PDF are converted to Word documents using OCR and then parsed into a structured JSON format. The JSON schema contains attributes of the question, choices, answer, domain category, and cognitive category, and there are four question types, including single-choice, multiple-choice, true/false, and open-ended Q\&A. For samples involving complex mathematical notation, expressions are manually converted into standard \LaTeX{} format following the conventions of C-Eval \cite{huang2024c} and MMLU \cite{hendrycks2020measuring}.

Each sample is initially categorized by agricultural experts using a custom annotation tool we developed.\footnote{http://www.widiagnosis.com:10008/problems} To ensure data quality, all entries are reviewed and corrected by expert annotators. To validate label consistency, we randomly sample 5\% of the data and ask two experts to independently annotate the questions, answers, and labels. Inter-annotator agreement is computed, and disagreements are resolved through discussion and targeted retraining until consistency exceeds 90\%. Only after reaching this threshold do we proceed with large-scale annotation. The Cohen's Kappa \cite{cohen1960coefficient} score between expert annotators for category labeling is 0.85. Consistency for question/option correctness and answer correctness reaches 99.7\%. The complete data processing pipeline is shown in Figure~\ref{fig-app-data-process}, with additional implementation details provided in Appendix~\ref{app_data_annotation}.

\subsection{Data Difficulty Enhancement}\label{s3-difficulty}
To better assess the model's discriminative ability to distinguish between options and enhance the distractiveness of the choices, we follow the practice of C-Eval \cite{huang2024c} and use GPT-4 to generate high-quality distractors. Each single-answer multiple-choice question is extended to include seven options, and all additional distractors are manually reviewed and validated by agricultural experts. This enhancement increases task difficulty while ensuring domain plausibility and consistency.

\subsection{Data Statistics}
\textbf{Basic Statistics.}  
AgriEval contains 14,697 multiple-choice questions and 2,167 open-ended Q\&A items, covering 29 agricultural subfields and 15 cognitive categories. Each subfield includes at least 100 questions, while each cognitive category contains over 2,000 samples. The average question length is 76.92 tokens, and the average answer length for generation tasks is 467.30 tokens. Table~\ref{tab:cognitive_data_statistics} and Table~\ref{tab-ap-domian-sta} show detailed distributions by domain and cognitive level, and Figure~\ref{fig:app_data_exp_generation} illustrates representative examples.

\textbf{Semantic Diversity.}  
To assess knowledge breadth, we visualize the semantic embedding space of AgriEval using BERT-based representations \cite{devlin-etal-2019-bert}. As shown in Figure~\ref{fig:semantic_diversity}, the embeddings demonstrate broad dispersion, indicating rich semantic coverage across domains. This suggests that AgriEval presents a diverse and challenging benchmark for LLMs.

\begin{table}[]
\centering
\caption{The performance of LLMs in zero-shot settings for cognitive tasks related to multi-choice questions. The best value within each model series is marked in bold, and the second-best is underlined.}
\label{tab:cognitive_main_result}
\resizebox{\textwidth}{!}{%
\begin{tabular}{cccccccccccccc}
\toprule
\multirow{2}{*}{\textbf{Model}} & \multicolumn{4}{c}{\textbf{Memorization}}                 & \multicolumn{3}{c}{\textbf{Understanding}} & \multicolumn{5}{c}{\textbf{Inference}}                                   & \multirow{2}{*}{\textbf{Overall}} \\ \cmidrule(lr){2-5} \cmidrule(lr){6-8} \cmidrule(lr){9-13}
                                & \textbf{M-P} & \textbf{M-R} & \textbf{M-E} & \textbf{M-T} & \textbf{U-I} & \textbf{U-S} & \textbf{U-V} & \textbf{I-D} & \textbf{I-N} & \textbf{I-P} & \textbf{I-S} & \textbf{I-G} &                                   \\ \midrule
Llama2-70B-Chat                 & 22.56        & 26.23        & 25.80         & 29.55        & 21.17        & 25.23        & 51.81        & 29.53        & 17.77        & 26.54        & 33.33        & 16.82        & 27.05                             \\
Mistral-7B-Instruct             & 24.44        & 21.31        & 36.10         & 37.12        & 21.96        & 29.25        & 48.84        & 34.99        & 21.28        & 25.90         & 45.18        & 21.18        & 29.10                              \\
Llama3-8B-Instruct              & 27.77        & 24.04        & 35.38        & 41.67        & 23.66        & 32.38        & 47.41        & 36.23        & 23.01        & 37.72        & 48.84        & 19.00           & 31.38                             \\
Baichuan2-7B-Chat               & 28.86        & 21.31        & 36.82        & 25.00           & 27.49        & 31.19        & 50.01        & 36.06        & 17.82        & 34.18        & 53.48        & 23.36        & 32.46                             \\
InternLM2-Chat-7B               & 29.49        & 22.95        & 38.79        & 31.06        & 27.44        & 32.78        & 49.24        & 43.92        & 21.49        & 40.48        & 59.46        & 21.81        & 33.58                             \\
DeepSeek-LLM-7B-Chat            & 29.76        & 22.95        & 39.28        & 31.06        & 29.29        & 32.10         & 51.45        & 39.70         & 17.14        & 39.00           & 50.43        & 24.61        & 33.76                             \\
Phi-3.5-Mini-Instruct           & 30.08        & 28.96        & 47.65        & 50.76        & 25.97        & 37.01        & 46.12        & 37.63        & 25.79        & 47.42        & 47.13        & 23.05        & 34.22                             \\
Mistral-Nemo-Instruct           & 30.2         & 25.68        & 37.8         & 50.76        & 29.99        & 36.28        & 50.01        & 37.55        & 21.49        & 39.35        & 52.99        & 23.99        & 34.39                             \\
Baichuan2-13B-Chat              & 32.27        & 32.24        & 46.10         & 33.33        & 28.78        & 34.89        & 47.27        & 44.50         & 24.21        & 45.51        & 49.69        & 25.55        & 35.53                             \\
ChatGLM3-6b                     & 32.04        & 29.51        & 43.26        & 38.64        & 30.77        & 36.25        & 49.70         & 44.58        & 21.80         & 39.49        & 42.37        & 26.79        & 35.55                             \\
Phi-3-Medium-4k-Instruct        & 33.14        & 34.43        & 44.73        & 52.27        & 28.23        & 42.09        & 49.86        & 40.36        & 26.21        & 48.48        & 49.45        & 25.86        & 36.87                             \\
Marco-o1                        & 34.77        & 33.33        & 35.91        & 36.36        & 33.66        & 36.43        & 49.74        & 43.51        & 27.62        & 35.46        & 66.06        & 28.04        & 37.32                             \\
Qwen2.5-3B-Instruct             & 37.14        & 36.61        & 43.07        & 54.55        & 33.33        & 39.60         & 50.65        & 42.51        & 37.58        & 39.70         & 61.66        & 29.28        & 39.67                             \\
Yi-1.5-9B-Chat                  & 37.46        & 24.04        & 44.85        & 40.15        & 36.47        & 42.72        & 58.66        & 38.30         & 30.50         & 43.74        & 58.00           & 16.51        & 41.20                              \\
InternLM2.5-20b-Chat            & 39.80         & 40.44        & 42.23        & 39.39        & 35.98        & 42.50         & 53.90         & 41.85        & 43.55        & 44.02        & 64.22        & 29.60         & 42.26                             \\
Llama3-70B-Instruct             & 40.45        & 36.07        & 45.11        & 45.45        & 37.24        & 49.09        & 49.41        & 45.08        & 39.05        & 56.05        & 62.27        & 32.09        & 43.16                             \\
GLM-4-9b-Chat                   & 41.85        & 43.72        & 51.48        & 50.00           & 37.14        & 48.69        & 51.56        & 46.48        & 22.69        & 59.24        & 58.24        & 26.17        & 43.72                             \\
InternLM2.5-7B-Chat             & 41.51        & 39.34        & 46.40         & 48.48        & 38.75        & 44.36        & 52.71        & 47.39        & 36.06        & 51.52        & 65.08        & 25.23        & 43.75                             \\
KwooLa                          & 43.51        & 33.88        & 49.51        & 57.58        & 42.18        & 45.49        & 49.72        & 46.65        & 24.90         & 52.30        & 60.56        & 31.46        & 44.48                             \\
Qwen2.5-7B-Instruct             & 46.4         & 48.09        & 53.14        & 59.85        & 45.21        & 49.97        & 53.68        & 50.54        & 36.90         & 54.49        & 66.18        & 31.78        & 48.21                             \\
Qwen2-7B-Instruct               & 47.79        & 43.72        & 49.39        & 53.79        & 47.86        & 50.48        & 56.26        & 46.32        & 28.67        & 53.43        & 68.01        & 26.17        & 48.83                             \\
Qwen2.5-14B-Instruct            & 47.81        & 47.54        & 47.05        & 56.82        & 47.68        & 51.64        & 55.74        & 48.80         & 49.21        & 44.44        & 72.04        & 33.96        & 49.53                             \\
Yi-1.5-34b-Chat                 & 50.24        & 45.36        & 51.44        & \underline{66.67}        & 50.57        & 52.67        & 62.28        & 46.82        & 36.32        & 53.86        & 66.91        & 30.22        & 51.83                             \\
Qwen2.5-32B-Instruct            & 55.32        & 53.55        & 52.05        & 65.91        & 54.31        & 57.80         & 61.53        & 55.17        & \underline{55.19}        & 55.34        & 73.50         & 50.16        & 56.35                             \\
DeepSeek-V3                     & 56.39        & 54.10         & 57.05        & 59.09        & 52.06        & 58.61        & 59.66        & \textbf{61.29}        & \textbf{65.25}        & \underline{61.15}        & \textbf{76.19}        & \textbf{53.27}        & 57.43                             \\
Qwen2.5-72B-Instruct            & \underline{60.15}        & \underline{56.83}        & \textbf{58.48}        & \textbf{70.45}        & \underline{60.45}        & \underline{61.91}        & \underline{62.69}        & 55.09        & 52.78        & 59.66        & 73.02        & \underline{51.09}        & \underline{60.32}                             \\
Qwen2-72B-Instruct              & \textbf{62.61}        & \textbf{57.92}        & \underline{58.11}        & \underline{66.67}        & \textbf{63.23}        & \textbf{65.26}        & \textbf{68.67}        & \underline{56.16}        & 45.65        & \textbf{63.55}        & \underline{73.63}        & \underline{51.09}        & \textbf{62.72}                             \\ \hdashline
GPT-3.5-Turbo                   & 31.20         & 31.15        & 39.55        & 36.36        & 28.67        & 36.10         & 49.16        & 40.45        & 19.34        & 35.67        & 52.75        & 22.43        & 34.43                             \\
GLM-4-Flash                     & 43.90         & 47.54        & 53.30         & \underline{59.09}        & 38.97        & 52.27        & 51.56        & 47.15        & 23.27        & \underline{62.63}        & 58.61        & 30.84        & 45.54                             \\
GPT-4o-mini                     & 46.98        & 45.90         & \textbf{59.89}        & 54.55        & 41.99        & \underline{56.04}        & 48.04        & \underline{56.82}        & 29.09        & 61.57        & 63.37        & 39.25        & 48.19                             \\
GPT-4o                          & 47.38        & 44.26        & 47.16        & 45.45        & 47.09        & 51.81        & 59.71        & 55.33        & 42.14        & 53.29        & \textbf{73.26}        & 42.99        & 50.01                             \\
GLM-4-Air                       & 48.07        & 44.26        & 53.30         & 54.55        & 47.98        & 53.93        & 56.91        & 55.58        & 30.66        & 55.84        & 68.13        & 28.97        & 50.05                             \\
Claude-3.5-Sonnet               & 52.49        & 50.82        & \underline{59.09}        & 50.00           & 50.51        & 55.97        & 61.45        & \textbf{59.80}         & 47.33        & \textbf{69.43}       & 67.77        & \underline{53.27}        & 54.92                             \\
Gemini-2.0-Flash                & 52.30         & \underline{57.38}        & 55.68        & 54.55        & 52.95        & 52.72        & 62.21        & 55.58        & \textbf{66.51}        & 60.51        & 71.06        & 52.34        & 55.33                             \\
Qwen-Turbo                      & \underline{54.85}        & 50.82        & 51.36        & \underline{59.09}        & \underline{55.04}        & 55.06        & \underline{64.46}        & 55.33        & 46.54        & 53.08        & 71.06        & 39.25        & \underline{55.76}                             \\
Qwen-Plus                       & \textbf{63.83}        & \textbf{60.66}        & 58.30         & \textbf{68.18}        & \textbf{63.78}        & \textbf{63.82 }       & \textbf{67.21}        & 53.85        & \underline{54.09}        & 59.24        & \textbf{73.26}        & \textbf{56.07}        & \textbf{63.21}                             \\ \bottomrule
\end{tabular}

}
\end{table}

\section{Experiment}\label{s4}

\subsection{Experimental Setup}\label{s41}

We conduct a systematic evaluation of 51 LLMs, comprising nine commercial and 42 open-source models that differ in architecture, parameter size, and language alignment. Open-source models are deployed on 4$\times$NVIDIA H800 80GB GPUs via local inference, while commercial models are accessed through official APIs. All models are evaluated at a generation temperature of 1.0 and a maximum token length of 2048, with the results averaged over three independent runs to ensure stability.

To assess LLMs' knowledge understanding and reasoning capabilities in agricultural scenarios, we design multiple evaluation setups. We test three prompting strategies: \textbf{Zero-Shot Prompting}, where the model directly outputs an answer; \textbf{CoT Prompting}, which encourages step-by-step reasoning; and \textbf{Few-Shot Prompting}, which includes five in-context examples sampled from different subcategories. To examine the effect of external knowledge, we implement an \textbf{RAG} setup using Chinese Wikipedia as the retrieval corpus and evaluate it on a 1,000-sample subset. Additionally, to evaluate models' sensitivity to answer position, we randomly \textbf{shuffle multiple-choice options}, following prior work on positional bias in LLMs~\cite{zheng2023large,du2024causal,navigli2023biases}.

For evaluation metrics, we use accuracy for all choice-based questions. In multi-answer questions, predictions are counted as correct only if all correct options are selected exactly. For open-ended Q\&A tasks, we apply ROUGE-L~\cite{lin2004rouge} to measure generation quality, following the LexEval~\cite{li2024lexeval}. Further implementation details are provided in Appendix~\ref{app:exp}.

\begin{table}[]
\centering
\caption{The performance of LLMs in zero-shot settings for domain tasks related to multi-choice questions. The best value within each model series is marked bold, and the second-best is underlined.}
\label{tab:domain_main_results}
\resizebox{0.8\textwidth}{!}{%
\begin{tabular}{cccccccc}
\toprule
\textbf{Model}            & \textbf{PP} & \textbf{Aqua} & \textbf{GS} & \textbf{TCH}   & \textbf{Fore} & \textbf{AST}   & \textbf{Overall} \\ \midrule
Mistral-7B-Instruct       & 29.40            & 32.44       & 33.33         & 21.53 & 30.45    & 26.81 & 29.10   \\
Llama3-8B-Instruct        & 31.56            & 33.08       & 36.41         & 25.24 & 33.27    & 30.03 & 31.38   \\
Baichuan2-7B-Chat         & 32.71            & 32.18       & 28.97         & 26.01 & 33.66    & 32.27 & 32.46   \\
DeepSeek-LLM-7B-Chat      & 34.08            & 33.33       & 24.87         & 26.44 & 37.66    & 32.88 & 33.76   \\
Phi-3.5-Mini-Instruct     & 34.34            & 35.96       & 37.44         & 29.8  & 39.03    & 32.06 & 34.22   \\
Mistral-Nemo-Instruct     & 34.30            & 33.97       & 26.67         & 33.76 & 36.35    & 35.39 & 34.39   \\
Baichuan2-13B-Chat        & 35.62            & 37.88       & 33.59         & 31.96 & 38.57    & 34.13 & 35.53   \\
ChatGLM3-6B               & 35.89            & 31.03       & 30.26         & 29.29 & 39.88    & 35.19 & 35.55   \\
Phi-3-Medium-4k-Instruct  & 37.05            & 38.72       & 37.18         & 32.13 & 38.97    & 35.34 & 36.87   \\
Marco-o1                  & 38.24            & 33.97       & 36.15         & 31.18 & 31.30    & 35.28 & 37.32   \\
Qwen2.5-3B-Instruct       & 40.01            & 39.29       & 45.13         & 36.61 & 35.04    & 39.11 & 39.67   \\
Yi-1.5-9B-Chat            & 41.62            & 37.63       & 41.28         & 39.53 & 41.32    & 39.67 & 41.20   \\
InternLM2.5-20b-Chat      & 42.45            & 39.49       & 47.95         & 39.28 & 39.23    & 43.08 & 42.26   \\
Llama3-70B-Instruct       & 43.66            & 42.50       & 43.59         & 38.42 & 40.67    & 41.70 & 43.16   \\
GLM-4-9B-Chat             & 44.27            & 41.41       & 40.26         & 42.12 & 44.60    & 40.86 & 43.72   \\
InternLM2.5-7B-Chat       & 43.91            & 41.15       & 47.95         & 45.48 & 39.69    & 43.95 & 43.75   \\
Qwen2.5-7B-Instruct       & 48.64            & 46.03       & 47.95         & 46.68 & 48.46    & 46.13 & 48.21   \\
Qwen2.5-14B-Instruct      & 50.14            & 46.54       & 52.82         & 48.41 & 43.55    & 47.97 & 49.53   \\
Yi-1.5-34b-Chat           & 52.60            & 45.45       & 46.15         & 48.49 & 52.32    & 49.39 & 51.83   \\
Qwen2.5-32B-Instruct      & 57.07            & 52.37       & 53.33         & 53.83 & 51.28    & 54.93 & 56.35   \\
DeepSeek-V3               & 57.74            & \underline{56.92}       & \textbf{68.46}         & 52.20  & 52.46    & 57.33 & 57.43   \\
Qwen2.5-72B-Instruct      & \underline{60.98}            & \textbf{57.05}       & \underline{65.13}         & \underline{57.02} & \underline{54.35}    & \underline{58.88} & \underline{60.32}   \\
Qwen2-72B-Instruct        & \textbf{63.49}            & 54.55       & 55.38         & \textbf{61.93} & \textbf{56.32}    & \textbf{62.71} & \textbf{62.72}   \\ \hdashline
GPT-3.5-Turbo             & 34.43            & 33.46       & 27.69         & 32.82 & 38.51    & 34.37 & 34.43   \\
GLM-4-Flash               & 46.10            & 42.69       & 41.54         & 43.67 & 46.95    & 42.74 & 45.54   \\
GPT-4o-mini               & 48.46            & 48.27       & 46.15         & 42.38 & 50.10    & 47.21 & 48.19   \\
GPT-4o                    & 51.04            & 46.54       & 48.46         & 41.34 & 42.63    & 48.31 & 50.01   \\
GLM-4-Air                 & 50.40            & 46.92       & 40.00         & 49.10 & 51.08    & 49.22 & 50.05   \\
Claude-3.5-Sonnet         & 55.60            & 53.85       & 56.92         & 46.51 & 53.63    & 52.66 & 54.92   \\
Gemini-2.0-Flash             & 55.39            & \underline{54.42}       & \textbf{63.08}         & 46.77 & \underline{55.01}    & \underline{56.74} & 55.33   \\
Qwen-Turbo                & \underline{56.63}            & 48.08       & 56.92         & \underline{52.71} & 52.65    & 53.57 & \underline{55.76}   \\
Qwen-Plus                 & \textbf{63.97}            & \textbf{57.50}        & \underline{59.23}         & \textbf{61.50}  & \textbf{55.21}    & \textbf{62.91} & \textbf{63.21}   \\ \bottomrule
\end{tabular}
}
\end{table}

\subsection{Main Results}\label{s44}

In this section, we conduct a comprehensive comparison of various LLMs on the AgriEval benchmark. To present the results more clearly, we highlight selected representative outcomes in Table~\ref{tab:cognitive_main_result} and Table~\ref{tab:domain_main_results}, while the complete experimental results for all models are provided in Appendix~\ref{sec:appendix_supplementary_results}. Based on these results, we summarize the key findings as follows:

\textbf{\textit{AgriEval remains a highly challenging benchmark.}}
LLMs achieve an average accuracy of 41.27\% on AgriEval, with the vast majority of models failing to reach the 60\% threshold. Even GPT-4o struggles with tasks requiring specialized agricultural knowledge, reflecting limited domain adaptation and reasoning capabilities. These results highlight both the difficulty of AgriEval in capturing real-world agricultural challenges and the substantial gap that remains for current LLMs to achieve expert-level performance in agricultural applications.

\textbf{\textit{LLMs struggle with inference, revealing reasoning gaps.}}  
From a cognitive perspective, LLMs perform significantly worse on inference tasks, especially those requiring numerical reasoning and genetic inference, compared to memorization and understanding tasks. This gap reflects their reliance on surface-level pattern recognition rather than grounded or compositional reasoning. Their difficulty with multi-step logic and quantitative computation reveals a key limitation in current architectures, underscoring the need for structured reasoning, symbolic grounding, or tool-augmented approaches in complex, domain-specific scenarios.

\textbf{\textit{Open-source models have begun to surpass proprietary models.}}
The best-performing open-source LLM achieves 62.72\% accuracy on AgriEval, surpassing most proprietary models, including GPT-4o. This demonstrates that the application potential of open-source LLMs in agricultural tasks is steadily increasing. This trend highlights the increasing strategic value of open-source technologies in advancing intelligent agricultural systems, particularly in scenarios where transparency, customization, and cost-efficiency are critical.

\textbf{\textit{Optimal LLM performance remains below expert level.}}
To evaluate LLMs against human expertise, we construct an expert validation set by uniformly sampling 1,500 questions across all categories. Three agricultural experts with PhDs are recruited to annotate and answer the questions. As shown in Figure~\ref{fig:intro}(\textit{Right}), the experts achieve an average accuracy of 70.62\%, outperforming the best-performing LLM by 4.84\%. This gap highlights that, despite recent advances, LLMs still struggle with high-level reasoning and domain-specific knowledge in agricultural tasks.

Notably, expert performance is also imperfect. While each expert possesses deep knowledge in specific areas, accuracy declines on questions outside their core domains. This reveals a shared limitation for both humans and LLMs: difficulty in generalizing across the full breadth of agricultural knowledge.

\begin{figure*}[]
    \centering
   \includegraphics[width=\linewidth]{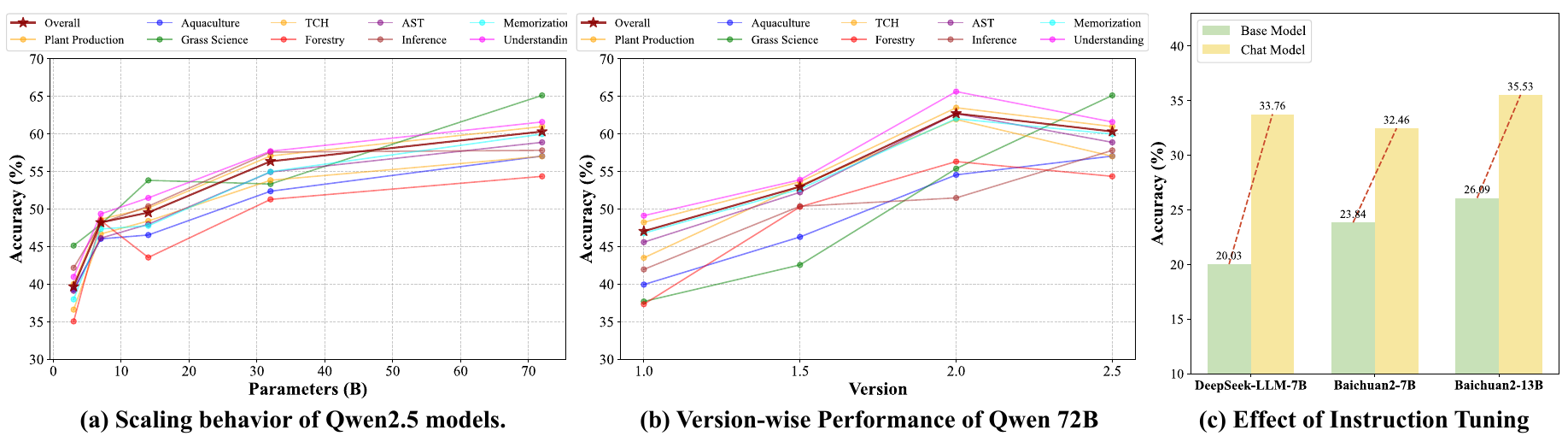}
    \vspace{-6mm}
    \caption{Performance trends across model scaling, version iteration, and instruction tuning.}
    \label{fig:further}
\end{figure*}

\subsection{Further Analysis}\label{s45}

\textbf{\textit{Larger models achieve better performance but exhibit diminishing returns.}} 
As shown in Figure \ref{fig:further}(a), we evaluate the performance of Qwen2.5 models across scales from 3B to 72B and observe that larger models generally achieve higher accuracy, aligning with the scaling law \cite{kaplan2020scaling}. However, the performance gains exhibit diminishing returns as the model size increases beyond 14B. This observation suggests that simply scaling up parameters is insufficient for solving complex domain-specific tasks, especially when domain adaptation or reasoning capability becomes the bottleneck.

\textbf{\textit{Version upgrade brings consistent gains.}}  
To isolate the effect of version iteration, we compare Qwen models of the same size (7B) across versions 1.0, 1.5, 2.0, and 2.5. As shown in Figure~\ref{fig:further}(b), each version upgrade brings consistent performance gains, likely due to improved pretraining, instruction tuning, and training data. Notably, Qwen2.5-7B significantly outperforms Qwen-7B, highlighting the importance of version optimization even at fixed model capacity.

\textbf{\textit{Instruction tuning significantly improves model performance and robustness.}} 
As shown in Figure~\ref{fig:further}(c), instruction-tuned models consistently outperform their base counterparts, with an average accuracy gain of 10.60\%. This improvement stems from supervised fine-tuning and alignment techniques that enhance instruction following and response quality. The performance gain is observed consistently across all question types, indicating stronger robustness in handling diverse task formats.

\textbf{\textit{Cross-lingual gaps challenge model generalization.}} 
Chinese-oriented LLMs perform moderately well on AgriEval, while English-oriented models like Llama \cite{touvron2023llama} consistently underperform. This reveals challenges in cross-lingual generalization, as English-pretrained models struggle with Chinese domain-specific content such as crop terms, regulatory language, and regional expressions. These results highlight the need for targeted pretraining or fine-tuning to bridge language gaps in non-English, high-stakes domains like agriculture.


\subsection{Exploration}\label{s46}
\begin{table}[]
\centering
\caption{Comparison of five models on multi-choice cognitive tasks under zero-shot, few-shot, and CoT settings. {\increase{}}/{\decrease{}} represents the performance increase/decrease compared to the zero-shot setting.}
\label{tab:exploration}
\resizebox{\linewidth}{!}{%
\begin{tabular}{ccccccccccccccc}
\toprule
\multirow{2}{*}{\textbf{Model}} & \multirow{2}{*}{\textbf{Prompt}} & \multicolumn{4}{c}{\textbf{Memorization}}                 & \multicolumn{3}{c}{\textbf{Understanding}} & \multicolumn{5}{c}{\textbf{Inference}}                          & \multirow{2}{*}{\textbf{Overall}} \\ \cmidrule(lr){3-6} \cmidrule(lr){7-9} \cmidrule(lr){10-14} 
    &       & \textbf{M-T} & \textbf{M-P} & \textbf{M-R} & \textbf{M-E} & \textbf{U-V} & \textbf{U-I} & \textbf{U-S} & \textbf{I-P} & \textbf{I-N} & \textbf{I-D} & \textbf{I-S}  & \textbf{I-G}  &                                   \\ \midrule
\multirow{3}{*}{Qwen2.5-3B-Instruct}  &    Zero-Shot   & 37.14        & 36.61        & 43.07        & 54.55        & 33.33        & 39.60         & 50.65        & 42.51        & 37.58        & 39.70         & 61.66        & 29.28        & 39.67             \\
                                      &    CoT       & 34.86\decrease{}        & 38.80\increase{}        & 41.48\decrease{}        & 39.39\decrease{}        & 30.98\decrease{}        & 39.85\increase{}        & 52.27\increase{}        & 36.06\decrease{}        & 49.16\increase{}        & 46.36\increase{}        & 58.36\decrease{}        & 24.30\decrease{}                            & 38.90\decrease{}     \\
                                      &    Few-Shot  & 38.48\increase{}         & 31.69\decrease{}         & 45.64\increase{}         & 37.88\decrease{}         & 35.21\increase{}         & 38.52\decrease{}         & 52.78\increase{}         & 41.77\decrease{}         & 33.81\decrease{}         & 42.25\increase{}         & 59.58\decrease{}         & 30.84\increase{}                             & 40.67\increase{}         \\  \midrule
\multirow{3}{*}{Qwen2.5-14B-Instruct}  &    Zero-Shot   & 47.81        & 47.54        & 47.05        & 56.82        & 47.68        & 51.64        & 55.74        & 48.80         & 49.21        & 44.44        & 72.04        & 33.96        & 49.53          \\
                                      &    CoT       & 45.31\decrease{}         & 40.44\decrease{}         & 44.77\decrease{}         & 41.67\decrease{}         & 44.37\decrease{}         & 48.92\decrease{}         & 57.84\increase{}         & 45.33\decrease{}         & 64.99\increase{}         & 47.13\increase{}         & 64.84\decrease{}         & 38.01\increase{}                             & 48.39\decrease{}        \\
                                      &    Few-Shot  & 51.57\increase{}         & 42.62\decrease{}         & 46.89\decrease{}         & 53.03\decrease{}         & 50.13\increase{}         & 53.17\increase{}         & 57.25\increase{}         & 50.87\increase{}         & 51.68\increase{}         & 48.20\increase{}         & 69.72\decrease{}         & 40.50\increase{}                             & 52.07\increase{}        \\  \midrule
\multirow{3}{*}{GLM-4-9B-Chat}  &    Zero-Shot  & 41.85        & 43.72        & 51.48        & 50.00           & 37.14        & 48.69        & 51.56        & 46.48        & 22.69        & 59.24        & 58.24        & 26.17        & 43.72         \\
                                      &    CoT         & 39.48\decrease{}         & 45.36\increase{}        & 47.65\decrease{}         & 44.70\decrease{}         & 35.80\decrease{}         & 48.09\decrease{}         & 52.92\increase{}        & 43.09\decrease{}         & 56.29\increase{}        & 56.97\decrease{}         & 60.68\increase{}        & 23.36\decrease{}                             & 43.78\increase{}       \\
                                      &    Few-Shot   & 38.25\decrease{}         & 42.62\decrease{}         & 42.12\decrease{}         & 53.03\increase{}        & 35.03\decrease{}         & 44.64\decrease{}         & 49.89\decrease{}         & 40.53\decrease{}         & 22.17\decrease{}         & 48.05\decrease{}         & 53.97\decrease{}         & 24.61\decrease{}                             & 40.07\decrease{}        \\  \midrule
\multirow{3}{*}{GPT-3.5-Turbo}  &    Zero-Shot    & 31.20         & 31.15        & 39.55        & 36.36        & 28.67        & 36.10         & 49.16        & 40.45        & 19.34        & 35.67        & 52.75        & 22.43        & 34.43          \\
                                      &    CoT         & 32.09\increase{}        & 40.98\increase{}        & 39.09\decrease{}        & 29.55\decrease{}        & 29.74\increase{}        & 40.18\increase{}        & 49.77\increase{}        & 34.00\decrease{}        & 49.69\increase{}        & 46.71\increase{}        & 47.99\decrease{}        & 17.76\decrease{}                            & 36.78\increase{}       \\
                                      &    Few-Shot     & 34.06\increase{}        & 45.90\increase{}        & 38.75\decrease{}        & 45.45\increase{}        & 30.54\increase{}        & 39.80\increase{}        & 49.62\increase{}        & 36.97\decrease{}        & 22.17\increase{}        & 36.09\increase{}        & 53.48\increase{}        & 29.91\increase{}                            & 36.47\increase{}      \\  \midrule
\multirow{3}{*}{GPT-4o-mini}  &    Zero-Shot  & 46.98        & 45.90         & 59.89        & 54.55        & 41.99        & 56.04        & 48.04        & 56.82        & 29.09        & 61.57        & 63.37        & 39.25        & 48.19          \\
                                      &    CoT        & 40.91\decrease{}        & 40.98\decrease{}        & 47.61\decrease{}        & 40.91\decrease{}        & 37.33\decrease{}        & 48.72\decrease{}        & 54.41\increase{}        & 30.77\decrease{}        & 41.19\increase{}        & 47.56\decrease{}        & 50.92\decrease{}        & 32.71\decrease{}                            & 43.29\decrease{}       \\
                                      &    Few-Shot   & 48.36\increase{}        & 52.46\increase{}        & 62.27\increase{}        & 72.73\increase{}        & 42.92\increase{}        & 57.78\increase{}        & 52.47\increase{}        & 58.31\increase{}        & 26.89\decrease{}        & 54.99\decrease{}        & 64.10\increase{}        & 36.45\decrease{}                            & 49.63\increase{}        \\  \bottomrule                                      
\end{tabular}
}
\end{table}

\textbf{\textit{CoT improves reasoning-intensive tasks but hinders performance on factual ones.}} 
To assess the effectiveness of CoT prompting, we compare model performance under zero-shot and CoT settings (Table~\ref{tab:exploration}). On average, CoT leads to a 3.51\% performance drop, aligned with MMLU~\cite{wang2024mmlu} and C-Eval~\cite{huang2024c}. This suggests that CoT may introduce unnecessary reasoning steps when shallow pattern matching or factual recall is sufficient. However, a fine-grained analysis reveals that CoT brings significant benefits in specific scenarios. For numerical reasoning tasks, CoT improves accuracy by 9.81\% on average, demonstrating its strength in guiding structured, multi-step computation. Moreover, on multi-answer multiple-choice questions, CoT enhances disambiguation and option filtering, leading to a 5.92\% accuracy gain (see Table~\ref{tab:appendix_choice_formats}). These improvements highlight CoT's potential in tasks that require step-wise reasoning or combinatorial decision-making. Taken together, these results suggest that the utility of CoT is highly task-dependent. Even though it may hinder performance on fact-based questions by introducing unnecessary complexity, it proves beneficial in inference-heavy contexts. Future prompting strategies may benefit from dynamic CoT selection mechanisms based on task type or reasoning difficulty.

\textbf{\textit{Few-shot learning cannot stably improve performance.}} 
We further explore the impact of in-context learning using a 5-shot setting, with results shown in Table~\ref{tab:exploration}. The results indicate that in-context learning yields inconsistent performance on AgriEval and does not always lead to improvements. We observe that model performance is highly sensitive to the relevance and quality of selected examples; semantically misaligned demonstrations may introduce noise and increase cognitive load. This suggests that in-context learning requires careful design in domain-specific tasks and that context effectiveness can be improved through semantically aligned example selection or demonstration filtering strategies.

\begin{figure}[ht]
    \centering
   \includegraphics[width=\linewidth]{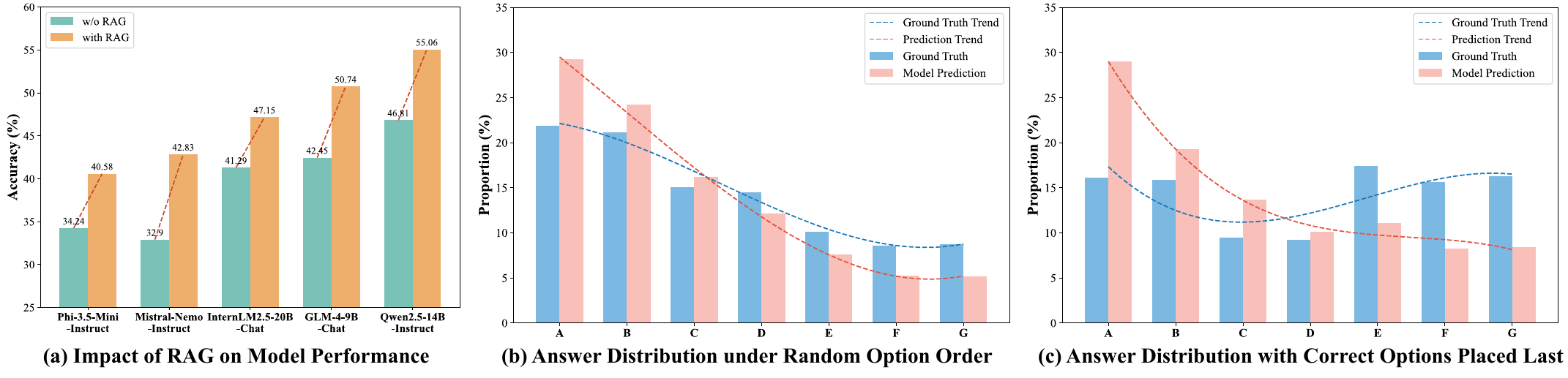}
    \vspace{-6mm}
    \caption{Exploratory analysis: effects of external knowledge and option order bias.}
    \label{fig:exploration}
\end{figure}

\textbf{\textit{RAG as an effective approach for rapid domain adaptation.}} 
To evaluate the impact of external knowledge, we construct a retrieval corpus from Chinese Wikipedia and randomly sample 1,000 examples across all categories to conduct RAG experiments. As shown in Figure~\ref{fig:exploration}(a), RAG consistently improves accuracy, with an average gain of approximately 4.0\%. Notably, smaller models benefit the most, suggesting that external knowledge can partially compensate for limited model capacity.
These results demonstrate the effectiveness of RAG in enhancing factual accuracy and knowledge grounding in agriculture domain tasks. However, varying performance gains across models highlight the need for better retrieval quality and more effective integration of retrieved information.

\textbf{\textit{LLMs exhibit positional bias in multiple-choice tasks.}} 
To assess LLMs' sensitivity to answer order, we conduct two experiments: (1) random shuffling option positions and (2) placing mostly correct answers in later positions (e.g., D–G). Results are shown in Figure~\ref{fig:appendix_shuffle}. Shuffling leads to an average 6.95\% accuracy drop, indicating reliance on positional cues. In the biased setting, although 58.50\% of correct answers appear later, models select them only 37.84\% of the time. As shown in Figure~\ref{fig:exploration}(b) and (c), predictions skewed toward earlier options, while ground-truth answers are more evenly distributed. These results align with prior studies \cite{zheng2023large,du2024causal,navigli2023biases}, which confirm that current LLMs favor positional heuristics over semantic reasoning. This calls for position-robust evaluation and training strategies, such as permutation augmentation and invariant prompting.

\subsection{Error Analysis}\label{s47}
\begin{wrapfigure}[19]{r}{0.45\textwidth} 
  \centering
  \includegraphics[width=0.4\textwidth]{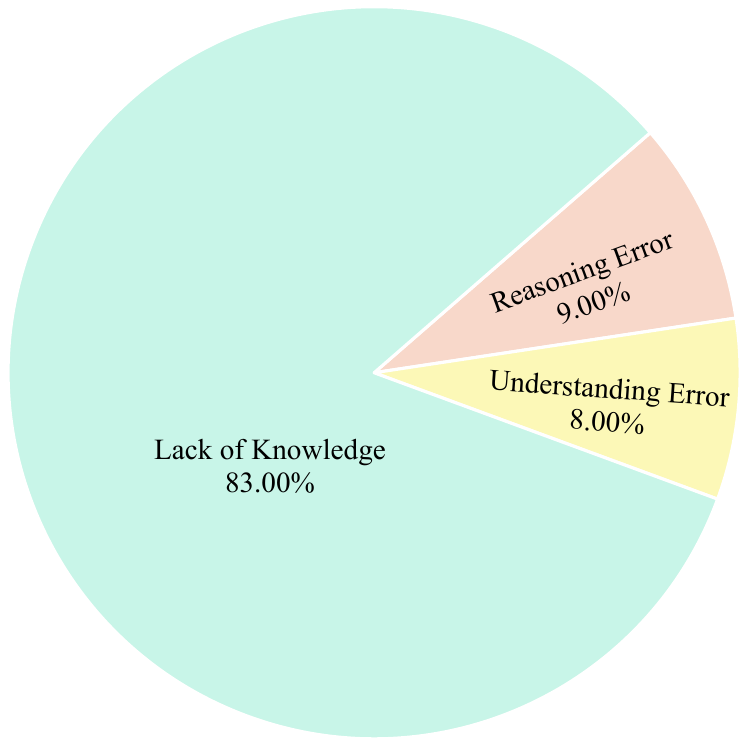}
  \caption{Overall error distribution for 200 annotated GPT-4o-mini errors.}
  \label{fig:main_error_analysis}
\end{wrapfigure}

In this section, we conduct an error analysis of GPT-4o-mini to uncover key limitations in domain-specific agricultural tasks and inform future improvements for LLM deployment. We sample 200 error cases and manually classify them into three categories: lack of knowledge, understanding error, and reasoning error. The distribution of error types is shown in Figure~\ref{fig:main_error_analysis}.

\vspace{4pt}

\vspace{-6pt}

\textbf{Lack of knowledge.} The majority of errors are caused by missing domain-specific knowledge. In these cases, the model fails to answer correctly due to insufficient domain-specific knowledge, particularly in agronomy, aquaculture, and forestry. For example, as shown in Figure~\ref{fig:knowledge-error}, the model fails to answer correctly due to missing information about lionfish parasites and groupers. This highlights the need for stronger domain grounding and specialized pretraining.

\textbf{Understanding errors.} These account for 8\% of cases and typically involve the model misinterpreting question intent or its own prior knowledge. For instance, in Figure~\ref{fig:understanding-error-1}, the model fails to identify the "most relevant" option despite all choices being contextually plausible. In Figure~\ref{fig:understanding-error-2}, although the model correctly explains the concept of protoplasts, it introduces inconsistency later in the explanation.

\textbf{Reasoning errors.} These are mostly found in numerical or procedural tasks involving biological quantities or resource planning. While CoT prompting improves step-by-step reasoning, the model still produces incorrect formulas or intermediate values, as illustrated in Figure~\ref{fig:reasoning-error}.

\section{Conclusion} \label{conclusion}
As the largest benchmark designed for agricultural production, AgriEval spans most agronomy subfields, aligns with human professional-level testing formats, and provides a comprehensive cognitive classification. This enables a systematic evaluation of current models' capabilities relative to human experts in agriculture. Our evaluation of 51 commercial and open-source LLMs reveals that even top-performing models, such as Qwen-Plus, struggle with real-world production management. Through an in-depth analysis examining factors such as model size, version differences, language orientation, the effectiveness of few-shot and CoT prompting, the necessity of external knowledge retrieval, generation bias, cognitive ability levels, and common errors, we identify key performance drivers and suggest areas for improvement. We believe AgriEval will assist smart agriculture developers in addressing knowledge gaps in agricultural LLMs, enhancing model capabilities, and providing insights for constructing benchmarks in other specialized fields.


\bibliographystyle{unsrtnat}
\bibliography{custom.bib}


\appendix
\clearpage
\section{Related Work}
\label{app:re}

\paragraph{Large Language Models}
ChatGPT \cite{ouyang2022training} demonstrates exceptional performance across various natural language processing tasks due to its advanced contextual understanding and language generation abilities. GPT-4 \cite{achiam2023gpt}, LLaMA \cite{touvron2023llama}, Qwen \cite{bai2023qwen,yang2024qwen2}, and other large language models (LLMs) have now widely permeated production and learning processes and even achieve human-like performance in knowledge question-and-answer (Q\&A) \cite{yang2023baichuan}, strategic planning \cite{wang2025learning}, coding \cite{wu2024internlm2}, and even complex reasoning \cite{liu2024deepseek}. Leveraging its powerful and efficient transfer capabilities, a large number of domain-specific LLMs have gradually emerged based on open-source general LLMs \cite{zhang2023huatuogpt,huang2023lawyer} and techniques like supervised fine-tuning (SFT) \cite{hu2021lora}, retrieval-augmented generation (RAG) \cite{lewis2020retrieval}, such as Bencao \cite{wang2023bencao}, Huatuo \cite{zhang2023huatuogpt}, Zhongjing \cite{yang2024zhongjing} in the medical, Lawyer \cite{huang2023lawyer} in law, and KwooLa in agriculture. Presently, building an intelligent, domain-specific brain centered around LLMs and creating autonomous domain application scenarios that integrate dialogue, reasoning, and tool usage has become one of the goals of artificial intelligence. However, since LLMs learn from vast amounts of data and make predictions through constructed network connections, they can easily generate illogical or factually incorrect decisions \cite{liu2024survey}. Directly applying them to costly, real-world production scenarios poses significant risks. Therefore, designing corresponding benchmarks to assess the level of human expertise achieved by LLMs is critical for their practical application and future development.

\section{Availability}\label{app:availability}
\begin{itemize}
    \item AgriEval dataset can be found at \url{https://huggingface.co/datasets/PaperHarvester/AgriEval}.
    \item The Github repository with evaluation code and prompts is available here: \url{https://github.com/YanPioneer/AgriEval/}.
    \item To enhance multilingual applicability and promote broader practical use, we translate all Chinese data into English using GPT-4o-mini. The translated dataset is publicly available at \url{https://huggingface.co/datasets/PaperHarvester/AgriEval}.
\end{itemize}

\section{Dataset AgriEval Details}\label{app:data}
\subsection{Task Overview} \label{app_task_overview}
AgriEval encompasses six primary agricultural domains and 29 subcategories derived from the structure of China's agricultural knowledge system. The benchmark is designed to assess four fundamental cognitive competencies: \textbf{Memorization}, \textbf{Understanding}, \textbf{Inference}, and \textbf{Generation}. These are further decomposed into 15 fine-grained skill dimensions to enable more precise evaluation. For clarity and reproducibility, we provide detailed definitions, task construction methodologies, and illustrative examples for each task type in the following sections.

\subsubsection{Domain Categories}
Guided by PhD-level agricultural experts, we define AgriEval's domain coverage based on the official classification system published by China's Ministry of Agriculture.\footnote{https://www.gov.cn/zhengce/zhengceku/2020-12/30/content\_5575377.htm} The benchmark includes practical and widely applicable agricultural scenarios spanning crop cultivation (e.g., staple, economic, and medicinal plants), forestry, livestock, aquaculture, and plant/animal protection. For subdomain alignment, we follow the disciplinary taxonomy adopted by China Agricultural University.\footnote{https://m.book118.com/html/2024/0915/7061025102006152.shtm} Given the dominant role of crop production and plant protection—accounting for approximately 54.9\% of China's total agricultural output\footnote{https://www.stats.gov.cn/sj/sjjd/202409/t20240912\_1956415.html?utm\_source=chatgpt.com}—the category is the most prominently represented in the dataset.

\subsubsection{Cognitive Ability Categories}
Building on Bloom's taxonomy~\cite{seaman2011bloom} and drawing inspiration from LexEval~\cite{li2024lexeval}, we organize the cognitive skills evaluated in AgriEval according to their alignment with real-world agricultural reasoning and decision-making. Each task is designed to reflect one or more levels of cognitive demand, ranging from factual memory to complex knowledge generation.

\begin{figure*}[htbp]
    \centering
    \makebox[\textwidth]{\includegraphics[width=\textwidth]{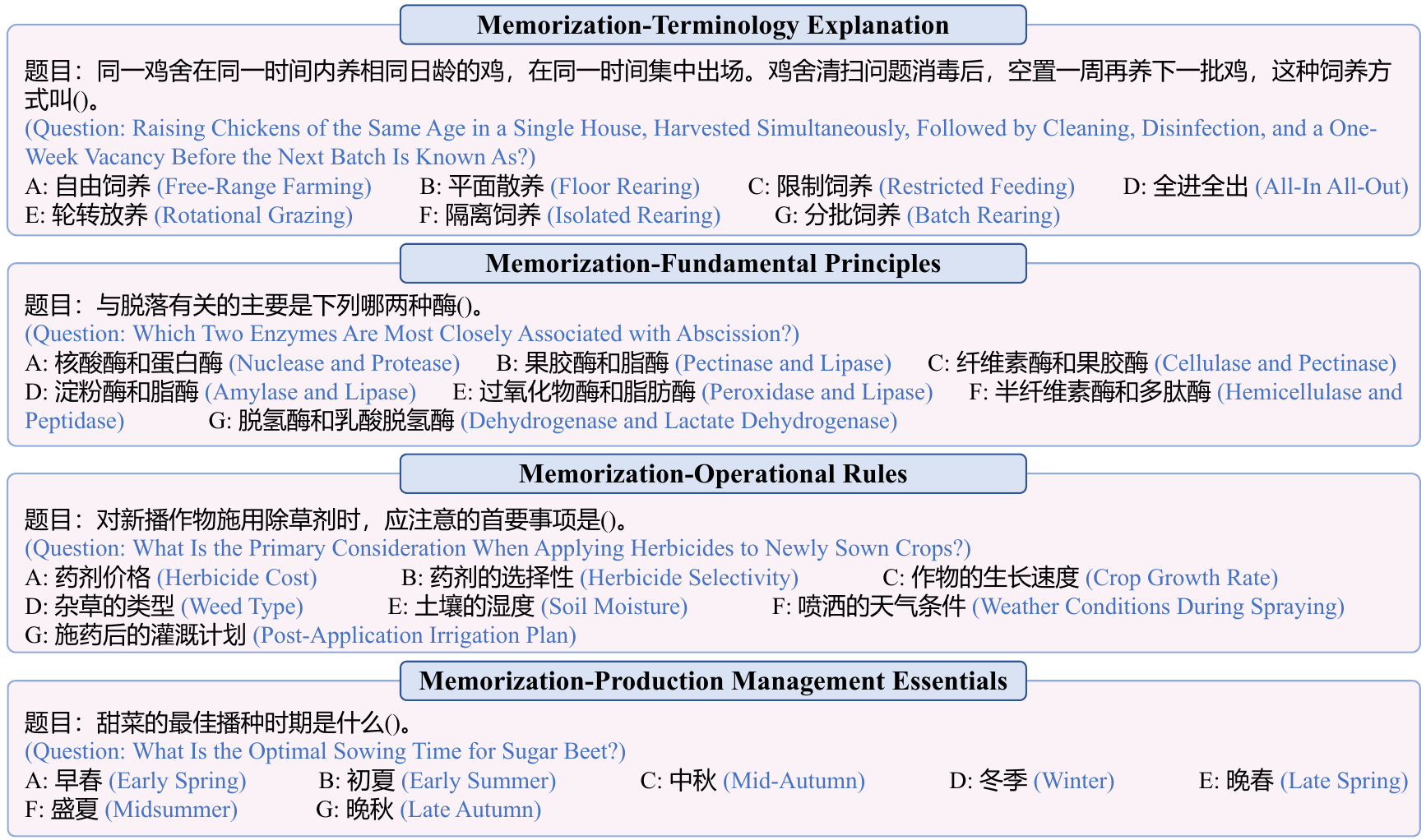}}
    \caption{Representative examples of memorization-level tasks, focusing on factual recall of agricultural terms, principles, and procedures.}
    \label{fig:app_data_exp_memorization}
\end{figure*}

\textbf{Memorization}: This category evaluates a model's capacity to recall and recognize factual agricultural knowledge, including terminology, core concepts, and standard procedures. Representative examples are illustrated in Figure~\ref{fig:app_data_exp_memorization}.

\begin{itemize}[leftmargin=2em]
    \item \textit{Terminology Explanation}: Recall and define domain-specific agricultural terms with accuracy.
    \item \textit{Fundamental Principles}: Recall foundational theories from plant physiology, soil chemistry, and general agronomic science.
    \item \textit{Operational Rules}: Recall standardized operating procedures and regulatory guidelines, such as pesticide usage instructions, safety intervals, and fertilizer application protocols.
    \item \textit{Production Management Essentials}: Recall critical technical practices and scheduling requirements for crop cultivation and livestock development.
\end{itemize}

\begin{figure*}[htbp]
    \centering
    \makebox[\textwidth]{\includegraphics[width=\textwidth]{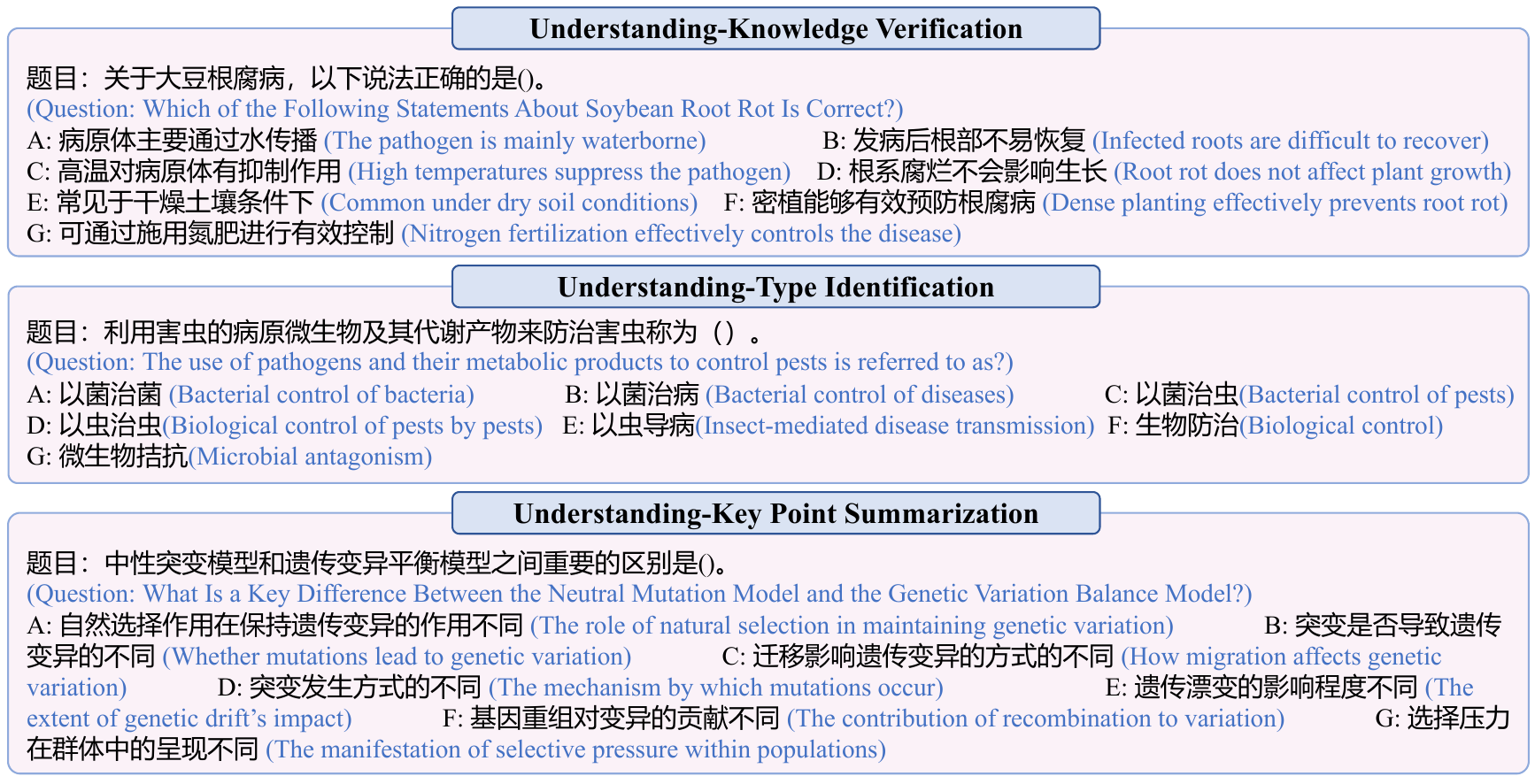}}
    \caption{Representative examples of understanding-level tasks involving knowledge verification, type identification, and key point summarization.}
    \label{fig:app_data_exp_understanding}
\end{figure*}

\textbf{Understanding}: This category assesses the model's ability to interpret agricultural concepts, identify key information, and distinguish between similar entities. Representative examples are provided in Figure~\ref{fig:app_data_exp_understanding}.

\begin{itemize}[leftmargin=2em]
    \item \textit{Knowledge Verification}: Assess the factual accuracy of given statements based on domain knowledge and scientific principles.
    \item \textit{Type Identification}: Identify agricultural entities by analyzing descriptive features such as morphology, structure, or composition.
    \item \textit{Key Point Summarization}: Extract and summarize hazards, impacts, or trends described in the input context.
\end{itemize}

\begin{figure*}[htbp]
    \centering
    \makebox[\textwidth]{\includegraphics[width=\textwidth]{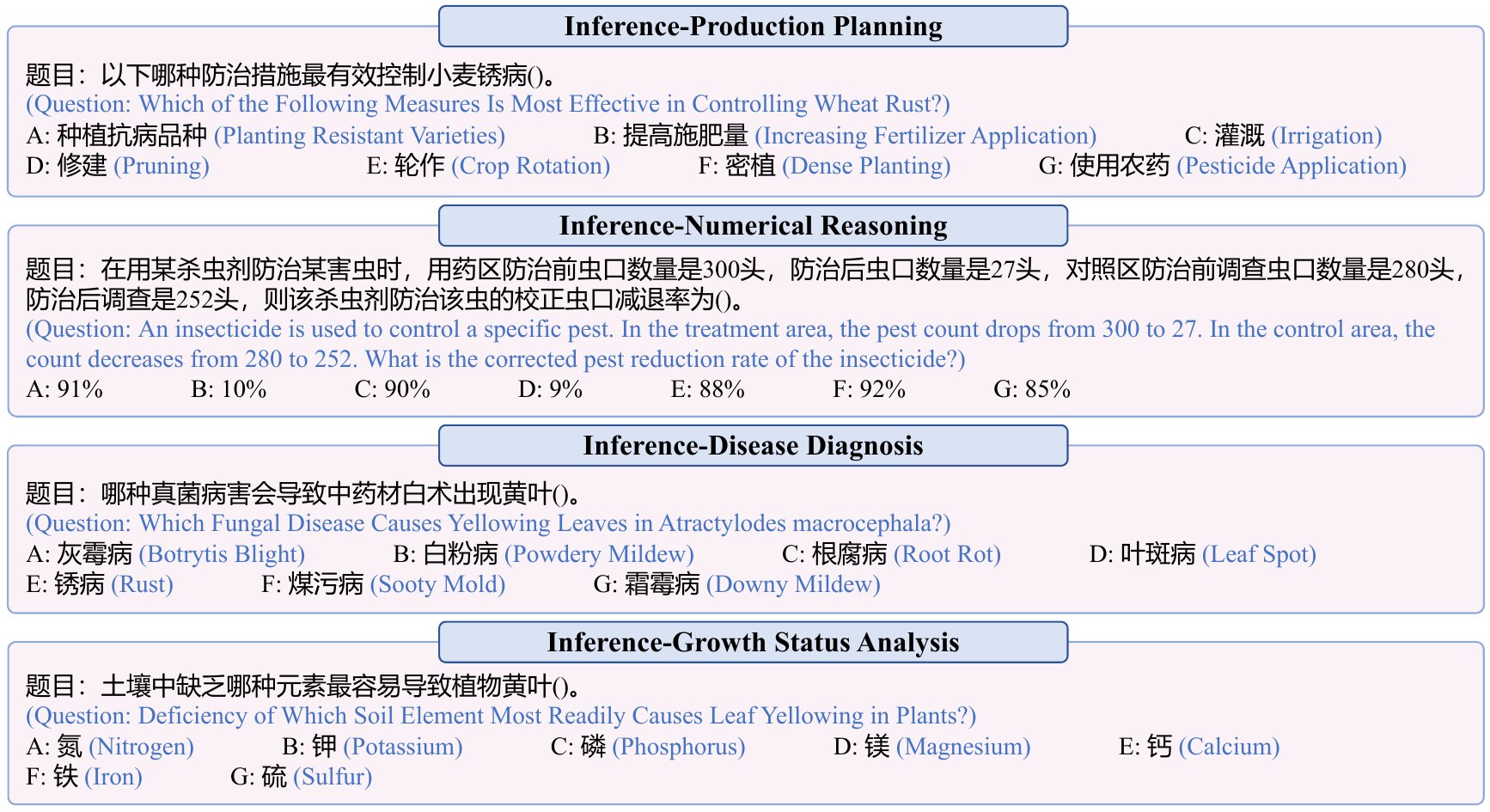}}
    \caption{Representative examples of inference-level tasks, including production planning, numerical reasoning, disease diagnosis, and growth status analysis.}
    \label{fig:app_data_exp_inference}
\end{figure*}

\textbf{Inference}: This category assesses the model's ability to perform reasoning grounded in agricultural knowledge, including cause analysis, decision recommendation, and quantitative estimation. Representative examples are shown in Figure~\ref{fig:app_data_exp_inference}.
\begin{itemize}[leftmargin=2em]
    \item \textit{Production Planning}: Recommend appropriate field management strategies based on environmental and operational constraints.
    \item \textit{Numerical Reasoning}: Compute key metrics such as seeding rate, fertilizer dosage, yield estimates, and irrigation requirements using domain-specific formulas or proportional reasoning.
    \item \textit{Disease Diagnosis}: Identify plant diseases and causal pathogens by interpreting symptoms such as leaf spots, chlorosis, or necrosis.
    \item \textit{Growth Status Analysis}: Analyze phenotypic indicators (e.g., leaf color, stem damage) to infer nutrient deficiencies, phytotoxic effects, or mechanical stress.
    \item \textit{Genetic Inference}: Infer the relationship between genotype and phenotype based on genetic principles, particularly the application of inheritance mechanisms such as dominance, recessiveness, and sex-linked traits.
\end{itemize}

\begin{figure*}[htbp]
    \centering
    \makebox[\textwidth]{\includegraphics[width=\textwidth]{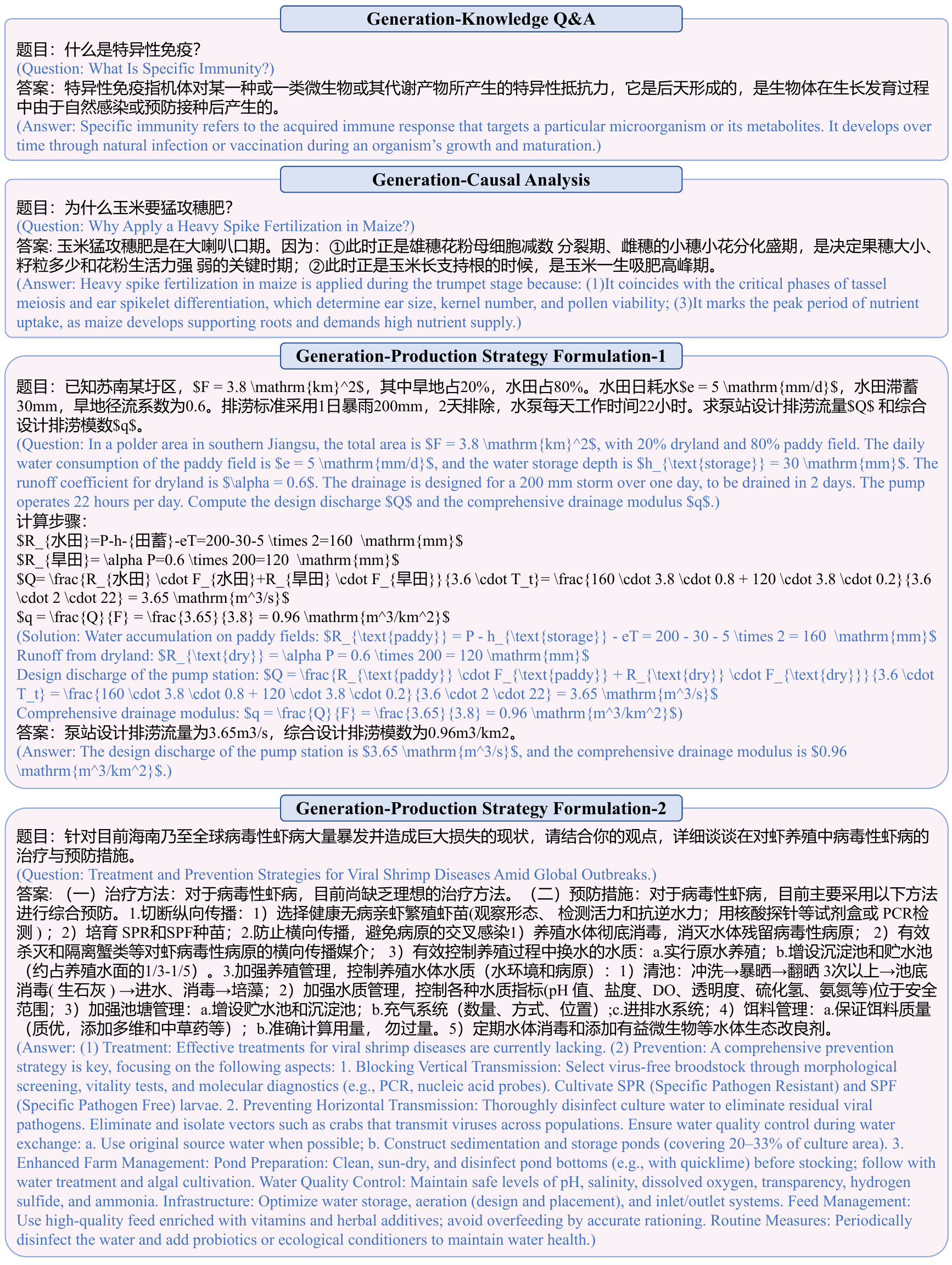}}
    \caption{Representative examples of generation-level tasks, involving knowledge Q\&A, causal analysis, and production strategy formulation.}
    \label{fig:app_data_exp_generation}
\end{figure*}

\textbf{Generation}: This category evaluates the model's ability to synthesize domain knowledge and generate coherent, contextually appropriate textual outputs. Tasks in this category require producing new strategies, explanations, or descriptive answers based on complex agricultural inputs. Representative examples are shown in Figure~\ref{fig:app_data_exp_generation}.

\begin{itemize}[leftmargin=2em]
    \item \textit{Knowledge Q\&A}: Generate detailed and actionable responses to questions involving theoretical concepts, operational procedures, or practical fieldwork.
    \item \textit{Production Strategy Formulation}: Integrate agronomic factors such as crop variety, fertilization, irrigation, and pest management into coherent and implementable production plans.
    \item \textit{Causal Analysis}: Generate plausible explanations for observed phenomena by attributing causes to environmental, managerial, or physiological factors.
\end{itemize}

\subsection{Data Collection and Licensing} \label{app_data_license}
We recruit two agricultural experts from partner university laboratories, both with advanced academic backgrounds. Prior to annotation, they are trained on the data collection objectives, scope (see Section~\ref{s3-data-collection}), annotation tools, and consistency protocols (see Section~\ref{s3-data-verification}). We begin with a clearly defined data collection guideline, requiring each annotator to gather an initial batch of 200 samples, followed by a large-scale collection only after quality checks are passed. The data sources fall into three main categories\footnote{https://m.book118.com}\footnote{https://easylearn.baidu.com/edu-page}\footnote{https://wenku.baidu.com/}:

\begin{enumerate}[leftmargin=2em]
    \item \textit{Public mock exam repositories}: Freely available question banks contributed by individual users and communities.
    \item \textit{Official postgraduate examination materials}: Publicly released by government institutions and licensed for public educational use.
    \item \textit{Archived university exams}: Questions collected and openly shared by students from top Chinese universities for academic purposes.
\end{enumerate}

All materials are originally in Word or PDF format. The experts manually search and collect over 500 documents. After filtering based on question difficulty, domain relevance, and alignment with real-world agricultural scenarios, 400 documents are retained for AgriEval. All data included in AgriEval complies with public usage and content-sharing policies. The dataset is released under the Apache License 2.0. Full licensing details are available at: \url{https://github.com/YanPioneer/AgriEval/}.

\begin{figure*}[htbp]
    \centering
    \makebox[\textwidth]{\includegraphics[width=\textwidth]{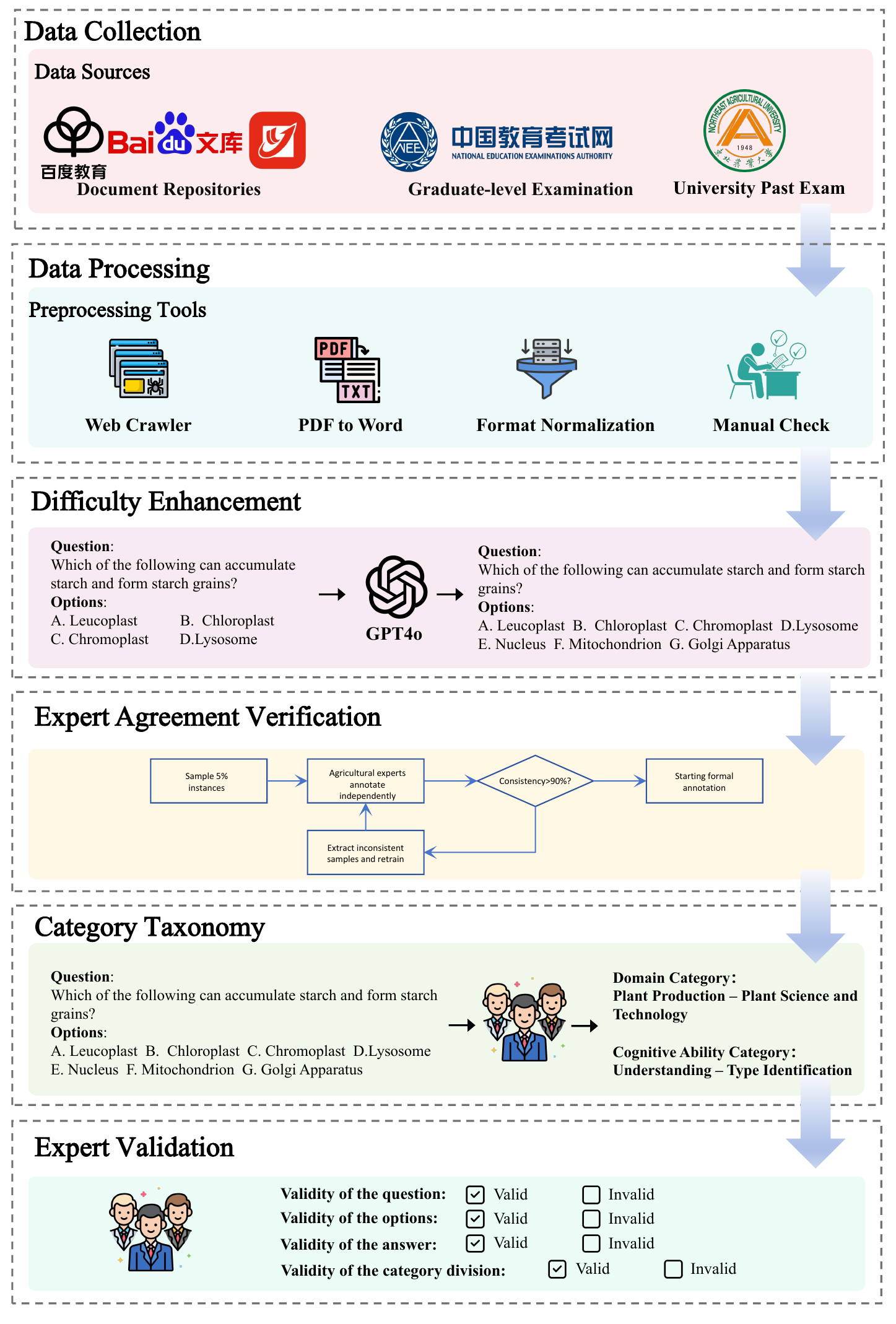}}
    \caption{The complete data processing pipeline.}
    \label{fig-app-data-process}
\end{figure*}

\begin{figure*}[htbp]
    \centering
    \makebox[\textwidth]{\includegraphics[width=\textwidth]{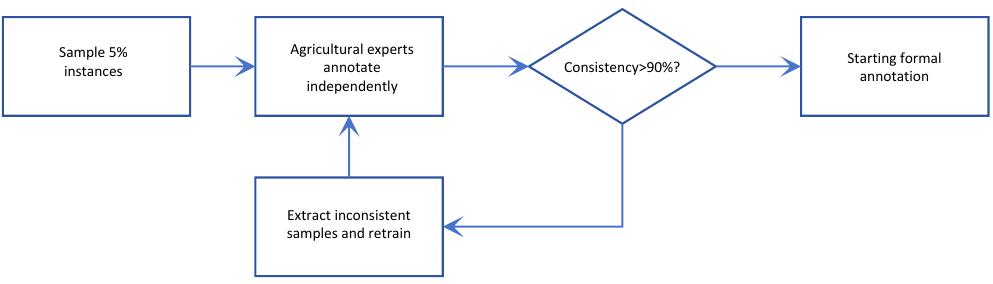}}
    \caption{Expert annotation training process. To ensure annotation quality and consistency, formal annotation begins only when expert agreement exceeds 90\%.}
    \label{fig-app-data-expert-consis}
\end{figure*}

\subsection{Data Annotation and Verification} \label{app_data_annotation}
All samples are verified and refined by domain experts in agriculture. To ensure annotation consistency across cognitive categories, we randomly sample 5\% of instances and have them independently annotated by two experts. Each annotation includes the question, candidate options, correct answer, and associated cognitive label.

Inter-annotator agreement is then assessed. If consistency falls below a 90\% threshold, targeted retraining is conducted using the inconsistent samples until agreement exceeds 90\%. Full-scale annotation is initiated only after the annotators reach this consistency benchmark. 

An example of the annotation format is shown in Figure~\ref{fig-app-data-process}, and the agreement verification workflow is illustrated in Figure~\ref{fig-app-data-expert-consis}.

\subsection{Data Statistics}

AgriEval consists of 14,697 multiple-choice questions and 2,167 open-ended Q\&A instances, spanning 29 domain sub-categories and 15 cognitive skill types relevant to agricultural applications. Table~\ref{tab:cognitive_data_statistics} summarizes the distribution of samples across cognitive categories, while Table~\ref{tab-ap-domian-sta} details their distribution across domain-specific categories.

To illustrate the semantic diversity of the dataset, Figure~\ref{fig:semantic_diversity} visualizes the embedding space of all questions using BERT~\cite{devlin-etal-2019-bert} representations projected via t-SNE~\cite{van2008visualizing}. Each color represents a distinct domain sub-category. The resulting distribution demonstrates AgriEval's broad semantic coverage and high inter-domain variability, supporting its utility as a comprehensive benchmark.

\begin{table}[]
\centering
\caption{AgriEval domain data statistics.}
\resizebox{\linewidth}{!}{%
\begin{tabular}{@{}clcc@{}}
\toprule
\textbf{Domain} & \multicolumn{1}{c}{\textbf{Sub-domain}} & \textbf{\#Samples} & \textbf{Avg. Tokens} \\ \midrule
 & \textbf{Plant Science and Technology} & \textbf{2523} & \textbf{74.38} \\
 & \textbf{Seed Science and Engineering} & \textbf{820}  & \textbf{57.46} \\
 & \textbf{Cultivation and Tillage} & \textbf{2392}  & \textbf{74.48} \\
 & \textbf{Plant Diseases} & \textbf{2183}  & \textbf{79.97} \\
 & \textbf{Weed Management} & \textbf{382}  & \textbf{83.22} \\
 & \textbf{Pest Management} & \textbf{1757}  & \textbf{78.64} \\
 & \textbf{Smart Agriculture} & \textbf{752}  & \textbf{76.75} \\
 & \textbf{Vegetables} & \textbf{249}  & \textbf{74.97} \\
 & \textbf{Fruiter} & \textbf{137}  & \textbf{71.93} \\
 & \textbf{Ecological Agriculture} & \textbf{335}  & \textbf{85.92} \\
 & \textbf{Cereal Crops} & \textbf{419}  & \textbf{70.32} \\
 & \textbf{Pesticides} & \textbf{224}  & \textbf{102.54} \\
 & \textbf{Oil Crops} & \textbf{157}  & \textbf{71.53} \\
 & \textbf{Cash Crops} & \textbf{147}  & \textbf{66.39} \\
 & \textbf{Tobacco} & \textbf{149}  & \textbf{126.13} \\
\multirow{-16}{*}{\textbf{Plant Production}}  & \textbf{Tea Science} & \textbf{204}  & \textbf{142.59} \\  \midrule
 & \textbf{Aquaculture Science} & \textbf{341}  & \textbf{72.62} \\
 & \textbf{Marine Science and Technology} & \textbf{114}  & \textbf{97.98} \\
 & \textbf{Aquatic Animal Medicine} & \textbf{191}  & \textbf{76.63} \\
\multirow{-4}{*}{\textbf{Aquaculture}} & \textbf{Aquarium Science and Technology} & \textbf{140}  & \textbf{69.37} \\ \midrule
\textbf{Grass Science} & \textbf{Grassland Science} & \textbf{218}  & \textbf{63.96} \\ \midrule
\textbf{Traditional Chinese Herbology} & \textbf{Cultivation and Identification of Chinese Herbs} & \textbf{388}  & \textbf{80.6} \\ \midrule
 & \textbf{Forest Protection} & \textbf{268}  & \textbf{94.19} \\
 & \textbf{Landscape Architecture} & \textbf{107}  & \textbf{99.7} \\
\multirow{-3}{*}{\textbf{Forestry}} & \textbf{Forestry Science} & \textbf{181}  & \textbf{77.17} \\ \midrule
 & \textbf{Feed   Engineering} & \textbf{130}  & \textbf{115.18} \\
 & \textbf{Husbandry Science} & \textbf{580}  & \textbf{62.4} \\
 & \textbf{Veterinary Medicine} & \textbf{774}  & \textbf{84.54} \\
\multirow{-4}{*}{\textbf{Animal Science and Technology}} & \textbf{Animal Science} & \textbf{602}  & \textbf{63.55} \\ \bottomrule
\end{tabular}%
}
\label{tab-ap-domian-sta}
\end{table}

\subsection{Broader Impact}\label{app-data-impact}
AgriEval aims to comprehensively evaluate the capabilities of large language models (LLMs) in real-world agricultural scenarios, promoting their responsible and reliable application in this high-stakes domain. We introduce a multi-domain, multi-level evaluation framework tailored to agriculture, offering researchers a structured tool to assess LLMs' cognitive abilities while providing standardized, comparable benchmarks for intelligent agriculture.

As a complex, knowledge-intensive field, agriculture demands accurate understanding, contextual reasoning, and multifactor decision-making. AgriEval emphasizes both the breadth and depth of agricultural knowledge, covering tasks from factual recall to complex reasoning. It helps uncover model strengths and limitations while guiding capability enhancement and deployment strategies. We believe this benchmark can accelerate the evolution of agricultural AI, shifting LLM development from general-purpose tools toward domain-specific competence and supporting developers, agronomists, and policymakers in making informed decisions.

Importantly, AgriEval does not imply that LLMs can replace agricultural experts or decision-makers. Agricultural decisions are grounded in long-term experience, local knowledge, and dynamic conditions, requiring continued human involvement. Our goal is to promote human-AI collaboration, not substitution, by revealing the boundaries and risks—such as hallucinations—of current models, thus laying the groundwork for sustainable, intelligent agricultural systems.

All data construction and usage in AgriEval follow rigorous ethical and fairness principles. We advocate for inclusive stakeholder participation to ensure diverse representation across crops, regions, and practitioners. We also call for ongoing evaluation and feedback loops in future LLM deployments to maximize societal value and ensure technical robustness.

\subsection{Ethical Considerations}\label{app-data-ethical}
AgriEval is reviewed for licensing compliance and data safety. All open-source materials used in the benchmark are properly licensed, with source details and licensing terms provided in the Appendix. All content is either publicly available or collected with proper authorization. We ensure that the dataset adheres to relevant legal and regulatory requirements and confirm that it is reviewed as part of the official filing process with the Cyberspace Administration of China.\footnote{https://www.cac.gov.cn/} To mitigate potential harm, we carefully filter the dataset to remove any content involving misinformation, regional discrimination, explicit or violent material, unfair competition, or offensive language. Based on internal review and domain expert assessment, we estimate that AgriEval poses minimal risk in terms of agricultural safety, fairness, regional bias, or other negative social impacts.

\section{Experiment Setup}\label{app:exp}
\subsection{Evaluated Prompts} \label{app-exp-prompt}

AgriEval includes three types of multiple-choice questions (single-answer, multiple-answer, and true/false), as well as open-ended Q\&A tasks. For multiple-choice questions, we evaluate LLM performance under three prompting strategies: zero-shot~\cite{romera2015embarrassingly}, few-shot~\cite{snell2017prototypical}, and CoT prompting~\cite{wei2022chain, chu-etal-2024-navigate}. The corresponding prompt templates are illustrated in Figure~\ref{fig:zero-normal-choice} (zero-shot), Figure~\ref{fig:few-normal-choice} (few-shot), and Figure~\ref{fig:zero-cot-choice} (CoT). For open-ended Q\&A tasks, we adopt the zero-shot setting to evaluate generalization without prior context. The prompt template used for generative Q\&A is shown in Figure~\ref{fig:zero-normal-qa}.

All prompt templates are originally written in Chinese, and the English versions presented in the figures are provided as translations for reference.

\begin{figure}[h]
\centering
\begin{prompt}[title={Prompt \thetcbcounter: Zero-Shot Normal for Multi-Choice}, label=choice_prompt_normal]
\begin{CJK}{UTF8}{gbsn}
以下是中国关于农业考试的单项选择题/多项选择题/判断题，请直接输出正确答案的选项，无需生成解释。\\
(The following are single-choice/multiple-choice/true-or-false questions for China's agricultural examination. Please directly input the correct answer option without generating an explanation.)\\

question：\{\textcolor{blue}{\textbf{question}}\} \\
(options：)\{\textcolor{blue}{\textbf{options\_str}}\}  \\
answer：
\end{CJK}
\end{prompt}
\caption{
The zero-shot prompt applied for multi-choice questions.
}
\label{fig:zero-normal-choice}
\end{figure}

\begin{figure}[htbp]
\centering
\begin{prompt}[title={Prompt \thetcbcounter: Few-Shot Normal for Multi-Choice}, label=choice_fewshot_prompt_normal]
\begin{CJK}{UTF8}{gbsn}
以下是中国关于农业考试的单项选择题/多项选择题/判断题，请直接输出正确答案的选项，无需生成解释。\\
(The following are single-choice/multiple-choice/true-or-false questions for China's agricultural examination. Please directly input the correct answer option without generating an explanation.)\\

以下是五个例子：\\
(Here are five examples:)\\
\{\textcolor{blue}{\textbf{examples}}\}\\
question：\{\textcolor{blue}{\textbf{question}}\} \\
(options：)\{\textcolor{blue}{\textbf{options\_str}}\}  \\
answer：
\end{CJK}
\end{prompt}
\caption{
The few-shot prompt applied for multi-choice questions.
}
\label{fig:few-normal-choice}
\end{figure}

\begin{figure}[htbp]
\centering
\begin{prompt}[title={Prompt \thetcbcounter: Zero-Shot CoT for Multi-Choice}, label=choice_cot_prompt_normal]
\begin{CJK}{UTF8}{gbsn}
以下是中国关于农业考试的单项选择题/多项选择题/判断题，回答时让我们一步步思考，逐个选项进行分析，最后输出答案。\\
(The following are single-choice questions/multiple-choice questions/true-or-false questions about China's agricultural examination. When answering, let us think step by step, analyze each option one by one, and finally output the answer.)\\

question：\{\textcolor{blue}{\textbf{question}}\} \\
(options：)\{\textcolor{blue}{\textbf{options\_str}}\}  \\
answer：
\end{CJK}
\end{prompt}
\caption{
The CoT prompt applied for multi-choice questions.
}
\label{fig:zero-cot-choice}
\end{figure}

\begin{figure}[htbp]
\centering
\begin{prompt}[title={Prompt \thetcbcounter: Zero-Shot Normal for Generation}, label=gen_prompt_normal]
\begin{CJK}{UTF8}{gbsn}
以下是中国关于农业考试的简答题，请输出正确答案。\\
(The following are questions and answers about China's agricultural examination. Please enter the correct answers.)\\

question：\{\textcolor{blue}{\textbf{question}}\} \\
answer：
\end{CJK}
\end{prompt}
\caption{
The zero-shot prompt applied for generation questions.
}
\label{fig:zero-normal-qa}
\end{figure}





\begin{table}[]
\centering
\caption{Summary of the 51 evaluated models on AgriEval, including nine proprietary and 42 open-source LLMs.}
\label{tab:appendix_models_statistics}
\resizebox{\textwidth}{!}{%
\begin{tabular}{ccccc}
\toprule
Model Type                    & Model & Size & Access & Parameter Link \\  \midrule
\multirow{43}{*}{Open-Source} & Baichuan2-7B-Base       &  7B     &    Weights    &   \url{https://huggingface.co/baichuan-inc/Baichuan2-7B-Base}             \\
                              & Baichuan2-7B-Chat       &  7B     &    Weights    &   \url{https://huggingface.co/baichuan-inc/Baichuan2-7B-Chat}             \\
                              & Baichuan2-13B-Base      &  13B    &    Weights    &   \url{https://huggingface.co/baichuan-inc/Baichuan2-13B-Base}            \\
                              & Baichuan2-13B-Chat      &  13B    &    Weights    &   \url{https://huggingface.co/baichuan-inc/Baichuan2-13B-Chat}            \\
                              & ChatGLM2-6B             &  6B     &    Weights    &   \url{https://huggingface.co/THUDM/chatglm2-6b}             \\
                              & ChatGLM3-6B             &  6B     &    Weights    &   \url{https://huggingface.co/THUDM/chatglm3-6b}             \\
                              & GLM-4-9B-Chat           &  9B     &    Weights    &   \url{https://huggingface.co/THUDM/glm-4-9b-chat}             \\
                              & DeepSeek-LLM-7B-Base    &  7B     &    Weights    &   \url{https://huggingface.co/deepseek-ai/deepseek-llm-7b-base}            \\
                              & DeepSeek-LLM-7B-Chat    &  7B     &    Weights    &   \url{https://huggingface.co/deepseek-ai/deepseek-llm-7b-chat}            \\
                              & DeepSeek-V3             &  671B(A37B)    &  API      &  \url{https://huggingface.co/deepseek-ai/DeepSeek-V3}              \\
                              & InternLM2-Chat-7B       &  7B     &    Weights    &   \url{https://huggingface.co/internlm/internlm2-chat-7b}             \\
                              & InternLM2.5-7B-Chat     &  7B     &    Weights    &   \url{https://huggingface.co/internlm/internlm2_5-7b-chat}              \\
                              & InternLM2.5-20B-Chat    &  20B    &    Weights    &   \url{https://huggingface.co/internlm/internlm2_5-20b-chat}             \\
                              & Llama-2-7b-chat-hf         & 7B         & Weights                     & \url{https://huggingface.co/meta-llama/Llama-2-7b-chat-hf}                    \\
                              & Llama-2-13b-chat-hf        & 13B        & Weights                     & \url{https://huggingface.co/meta-llama/Llama-2-13b-chat-hf}                   \\
                              & Llama-2-70b-chat-hf        & 70B        & Weights                     & \url{https://huggingface.co/meta-llama/Llama-2-70b-chat-hf}                   \\
                              & Meta-Llama-3-8B-Instruct   & 8B         & Weights                     & \url{https://huggingface.co/meta-llama/Meta-Llama-3-8B-Instruct}              \\
                              & Meta-Llama-3-70B-Instruct  & 70B        & Weights                     & \url{https://huggingface.co/meta-llama/Meta-Llama-3-70B-Instruct}             \\
                              & Marco-o1                   & 7.62B      & Weights                     & \url{https://huggingface.co/AIDC-AI/Marco-o1}                                 \\
                              & Mistral-7B-Instruct-v0.1   & 7B         & Weights                     & \url{https://huggingface.co/mistralai/Mistral-7B-Instruct-v0.1}               \\
                              & Mistral-Nemo-Instruct-2407 & 12.2B      & Weights                     & \url{https://huggingface.co/mistralai/Mistral-Nemo-Instruct-2407}             \\
                                                       & Phi-3-medium-4k-instruct   & 14B        & Weights                     & \url{https://huggingface.co/microsoft/Phi-3-medium-4k-instruct}               \\
                                                       & Phi-3.5-mini-instruct      & 3.82B      & Weights                     & \url{https://huggingface.co/microsoft/Phi-3.5-mini-instruct}                  \\
                                                       & Qwen-7B-Chat               & 7B         & Weights                     & \url{https://huggingface.co/Qwen/Qwen-7B-Chat}                                \\
                                                       & Qwen-14B-Chat              & 14B        & Weights                     & \url{https://huggingface.co/Qwen/Qwen-14B-Chat}                               \\
                                                       & Qwen-72B-Chat              & 70B        & Weights                     & \url{https://huggingface.co/Qwen/Qwen-72B-Chat}                               \\
                                                       & Qwen1.5-1.8B-Chat          & 1.8B       & Weights                     & \url{https://huggingface.co/Qwen/Qwen1.5-1.8B-Chat}                           \\
                                                       & Qwen1.5-4B-Chat            & 4B         & Weights                     & \url{https://huggingface.co/Qwen/Qwen1.5-4B-Chat}                             \\
                                                       & Qwen1.5-7B-Chat            & 7B         & Weights                     & \url{https://huggingface.co/Qwen/Qwen1.5-7B-Chat}                             \\
                                                       & Qwen1.5-14B-Chat           & 14B        & Weights                     & \url{https://huggingface.co/Qwen/Qwen1.5-14B-Chat}                            \\
                                                       & Qwen1.5-32B-Chat           & 32B        & Weights                     & \url{https://huggingface.co/Qwen/Qwen1.5-32B-Chat}                            \\
                                                       & Qwen1.5-72B-Chat           & 72B        & Weights                     & \url{https://huggingface.co/Qwen/Qwen1.5-72B-Chat}                            \\
                                                       & Qwen2-7B-Instruct          & 7B         & Weights                     & \url{https://huggingface.co/Qwen/Qwen2-7B-Instruct}                           \\
                                                       & Qwen2-72B-Instruct         & 72B        & Weights                     & \url{https://huggingface.co/Qwen/Qwen2-72B-Instruct}                          \\
                                                       & Qwen2.5-3B-Instruct        & 3B         & Weights                     & \url{https://huggingface.co/Qwen/Qwen2.5-3B-Instruct}                         \\
                                                       & Qwen2.5-7B-Instruct        & 7B         & Weights                     & \url{https://huggingface.co/Qwen/Qwen2.5-7B-Instruct}                         \\
                                                       & Qwen2.5-Coder-7B-Instruct  & 7B         & Weights                     & \url{https://huggingface.co/Qwen/Qwen2.5-Coder-7B-Instruct}                   \\
                                                       & Qwen2.5-14B-Instruct       & 14B        & Weights                     & \url{https://huggingface.co/Qwen/Qwen2.5-14B-Instruct}                        \\
                                                       & Qwen2.5-32B-Instruct       & 32B        & Weights                     & \url{https://huggingface.co/Qwen/Qwen2.5-32B-Instruct}                        \\
                                                       & Qwen2.5-72B-Instruct       & 72B        & Weights                     & \url{https://huggingface.co/Qwen/Qwen2.5-72B-Instruct}                        \\
                                                       & Yi-1.5-9B-Chat             & 9B         & Weights                     & \url{https://huggingface.co/01-ai/Yi-1.5-9B-Chat}                             \\
                                                       & Yi-1.5-34B-Chat            & 34B        & Weights                     & \url{https://huggingface.co/01-ai/Yi-1.5-34B-Chat}                            \\   \hdashline
\multirow{9}{*}{Proprietary}  & GLM-4-Flash            &  --    &  API      &   --             \\
                              & GLM-4-Air              &  --    &  API      &   --             \\
                              & Qwen-Trubo             &  --    &  API      &   --             \\
                              & Qwen-Plus              &  --    &  API      &   --             \\
                              & GPT-3.5-Turbo          &  --    &  API      &   --             \\
                              & GPT-4o-mini            &  --    &  API      &   --             \\
                              & GPT-4o                 &  --    &  API      &   --             \\
                              & Gemini-2.0-Pro         &  --    &  API      &   --             \\
                              & Claude-3.5-Sonnet      &  --    &  API      &   --             \\ \bottomrule
\end{tabular}
}
\end{table}

\subsection{Evaluated Models} \label{app-exp-model}
We evaluate a total of 51 LLMs on AgriEval, comprising nine proprietary and 42 open-source models. Detailed model configurations and parameter statistics are provided in Table~\ref{tab:appendix_models_statistics}. For open-source models, we download the corresponding weights and perform inference using the vLLM~\cite{kwon2023efficient} deployment API on 4$\times$NVIDIA H800 GPUs (80GB each). Proprietary models are evaluated via their official APIs.

\subsection{Evaluation Metrics} \label{app-exp-metric}

AgriEval includes both multiple-choice and open-ended Q\&A formats. For multiple-choice questions, we report accuracy by extracting the model's predicted options using regular expressions and comparing them with the ground truth labels. For open-ended Q\&A tasks, we adopt the Rouge-L score to evaluate the overlap between model-generated responses and reference answers, following the approach in~\cite{li2024lexeval}. 

The computation formulas for accuracy and Rouge-L are provided in Equation~\eqref{acc} and Equation~\eqref{rouge-l}, respectively.

\begin{equation}
        ACC=\frac{n}{N} \label{acc}
    \end{equation}
where $n$ denotes the amount of data answered correctly and $N$ denotes the total amount of all data.

\begin{equation}
        Rouge\_{L}=\frac{(1+ \alpha^2)R_{lcs}P_{lcs}}{R_{lcs}+\alpha^2 P_{lcs}} \label{rouge-l}
    \end{equation}
    where $\alpha=1$, $R_{lcs}= \frac{LCS(Y,A)}{Count(Y)}$, $P_{lcs}=\frac{LCS(Y,A)}{Count(A)}$, and $LCS(Y,A)$ represents the longest common subsequence between $Y$ and $A$.

\section{Supplementary Results} \label{sec:appendix_supplementary_results}

\begin{figure*}[htbp]
    \centering
    \makebox[\textwidth]{\includegraphics[width=1.3\textwidth]{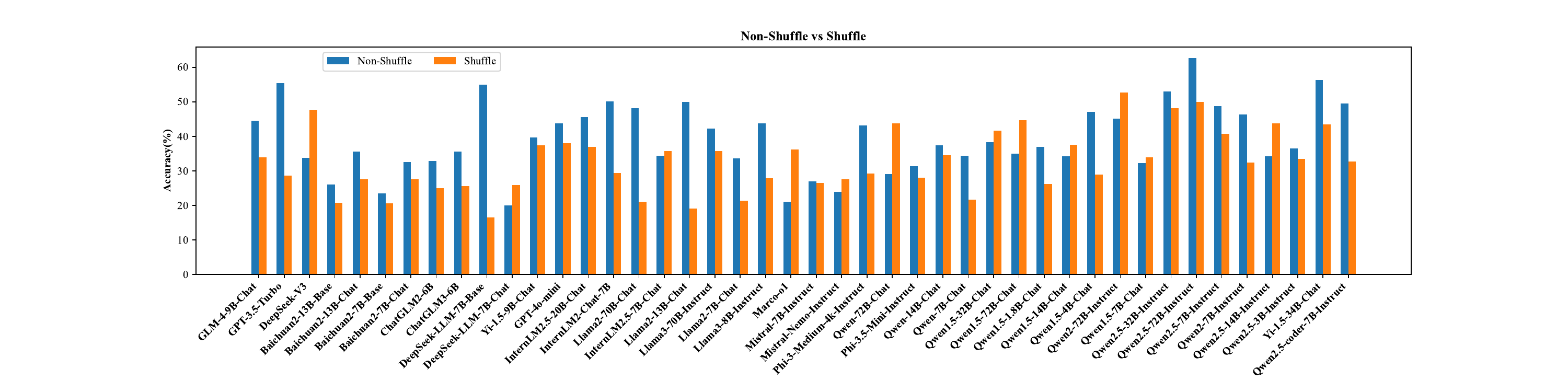}}
    \caption{Model performance after shuffling the order of options.}
    \label{fig:appendix_shuffle}
\end{figure*}

\begin{figure*}[htbp]
    \centering
    \makebox[\linewidth]{\includegraphics[width=\textwidth]{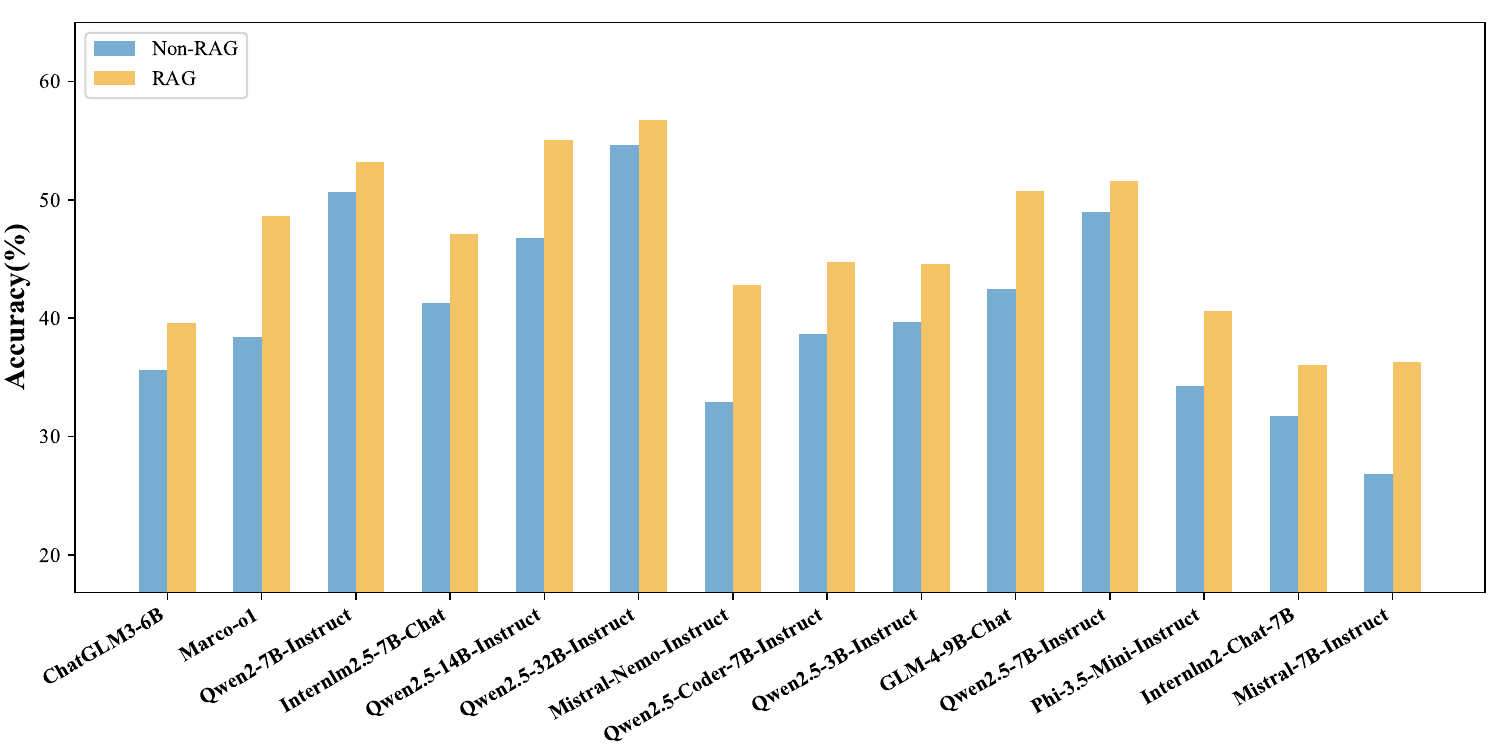}}
    \caption{Performance comparison with and without RAG across models.}
    \label{fig:appendix_rag}
\end{figure*}

In this section, we provide the full set of experimental results. Tables~\ref{tab:appendix_domain_choice_main_results_zero-shot_choice}–\ref{tab:appendix_domain_cot_choice_main_results} present model performance on cognitive and domain-specific multiple-choice tasks under zero-shot, CoT, and few-shot prompting settings. Table~\ref{tab:appendix_cognitive_domain_main_results_zero-shot_generation} reports zero-shot results on cognitive generation tasks. Figure~\ref{fig:appendix_shuffle} summarizes the performance variation under answer option shuffling across all models, and Figure~\ref{fig:appendix_rag} shows the impact of external knowledge augmentation using RAG. In addition, we present extended analyses to further support our experimental findings, as detailed below.

\begin{figure*}[htbp]
    \centering
    \makebox[\linewidth]{\includegraphics[width=\textwidth]{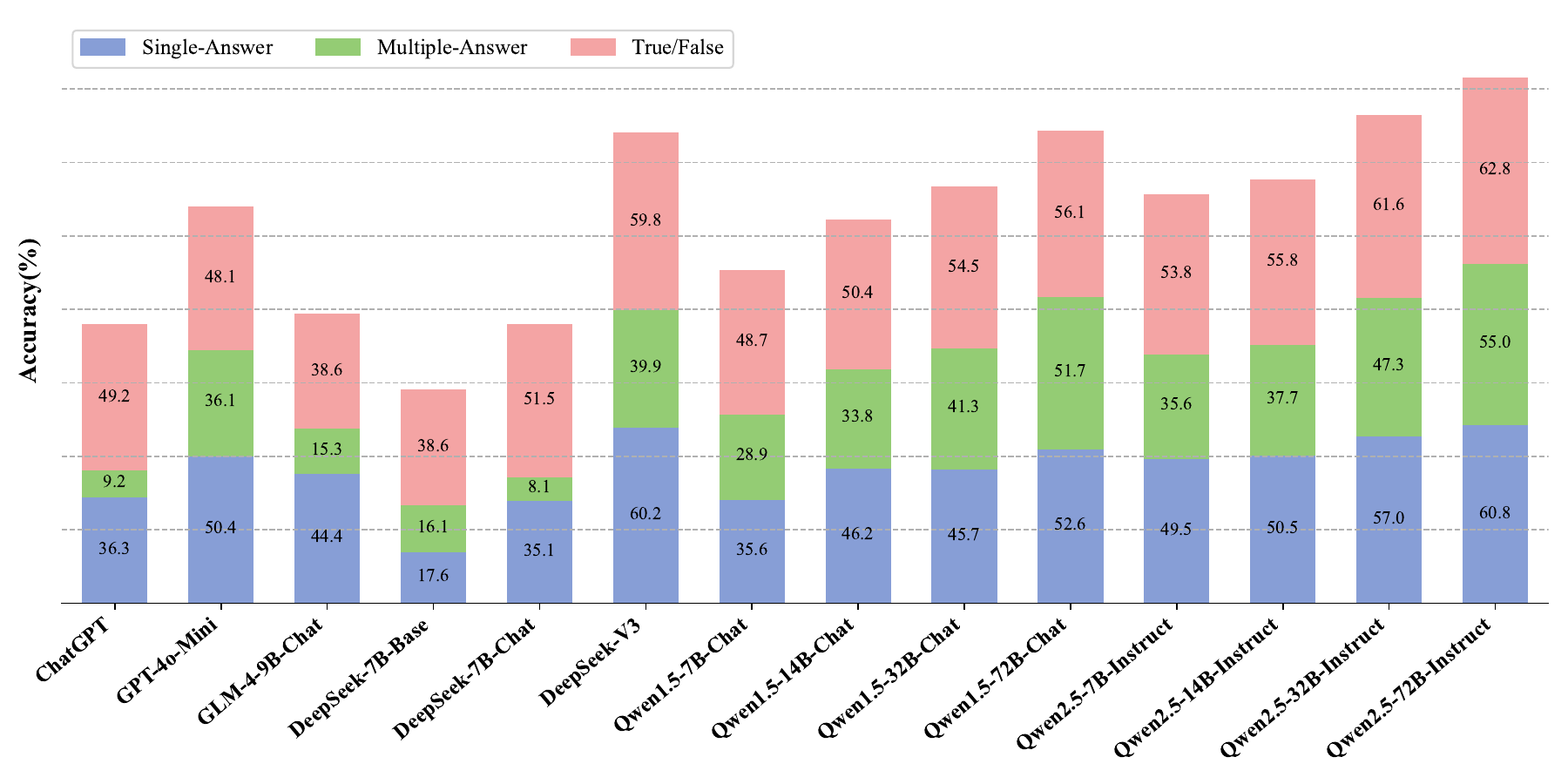}}
    \caption{Stacked bar chart of model performance across different multiple-choice formats, including single-answer, multi-answer, and true/false.}
    \label{fig:appendix_different_choice}
\end{figure*}

\textbf{Multi-answer multi-choice questions reveal LLMs' limitations in complex reasoning.}
We compare model performance across three multiple-choice formats: single-answer, multi-answer, and true/false, as shown in Figure~\ref{fig:appendix_different_choice}. Results indicate that multi-answer questions are the most challenging, while true/false questions yield the highest accuracy. The poor performance on multi-answer questions is likely due to their increased complexity, requiring the model to evaluate each option independently and make combinatorial judgments. This places higher demands on the model's reasoning ability, domain knowledge, and comprehensive analysis skills. In contrast, true/false questions are inherently simpler and can achieve up to 50\% accuracy through random guessing. These results highlight a key limitation of current LLMs: while they perform well on surface-level or binary classification tasks, they struggle with tasks that require deep, option-level reasoning and holistic knowledge integration.

\section{Error Analysis}\label{app-exp-error}
\begin{figure}[htbp]
    \centering
   \includegraphics[width=\linewidth]{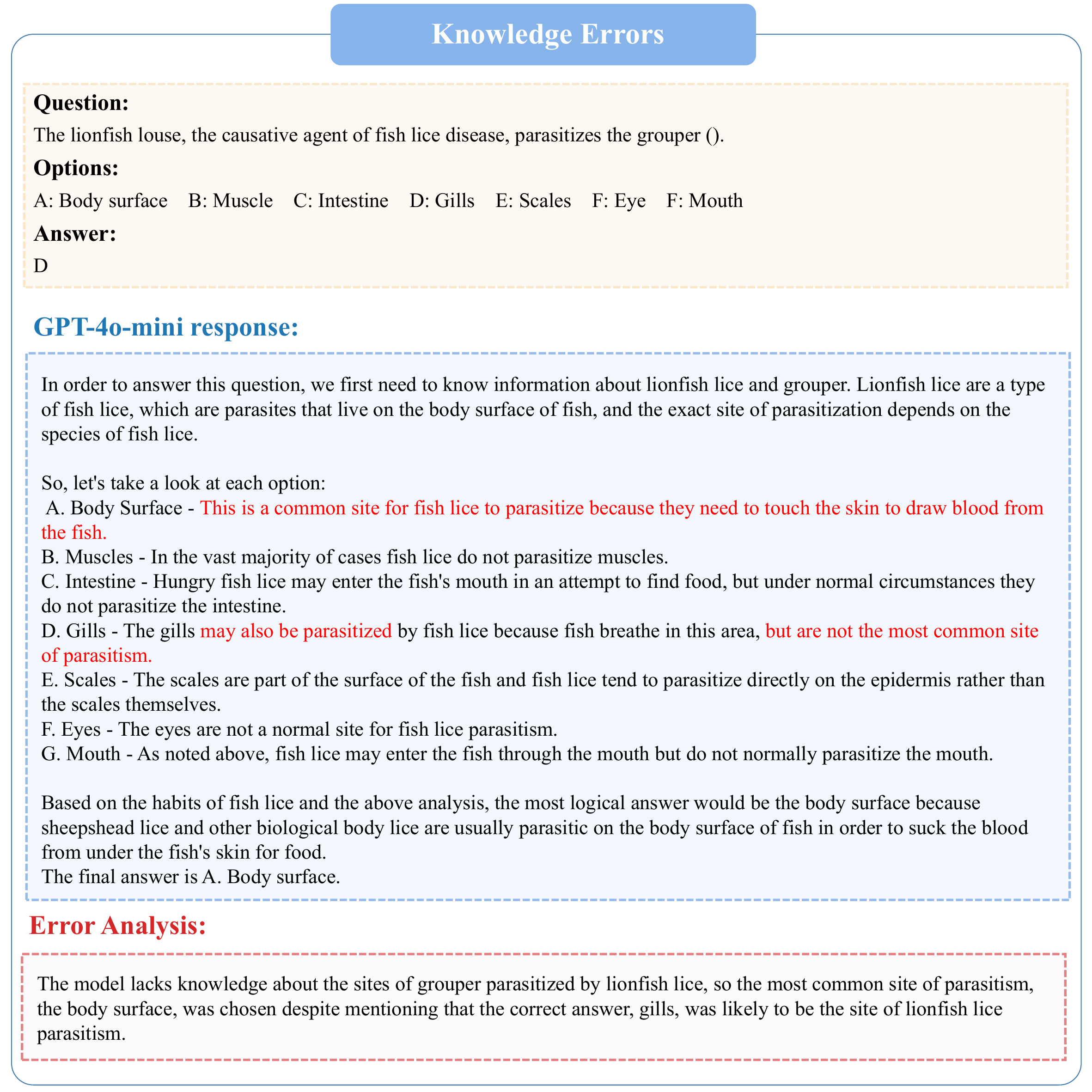}
    \caption{A sample for knowledge errors.}
    \label{fig:knowledge-error}
\end{figure}

\begin{figure}[htbp]
    \centering
   \includegraphics[width=\linewidth]{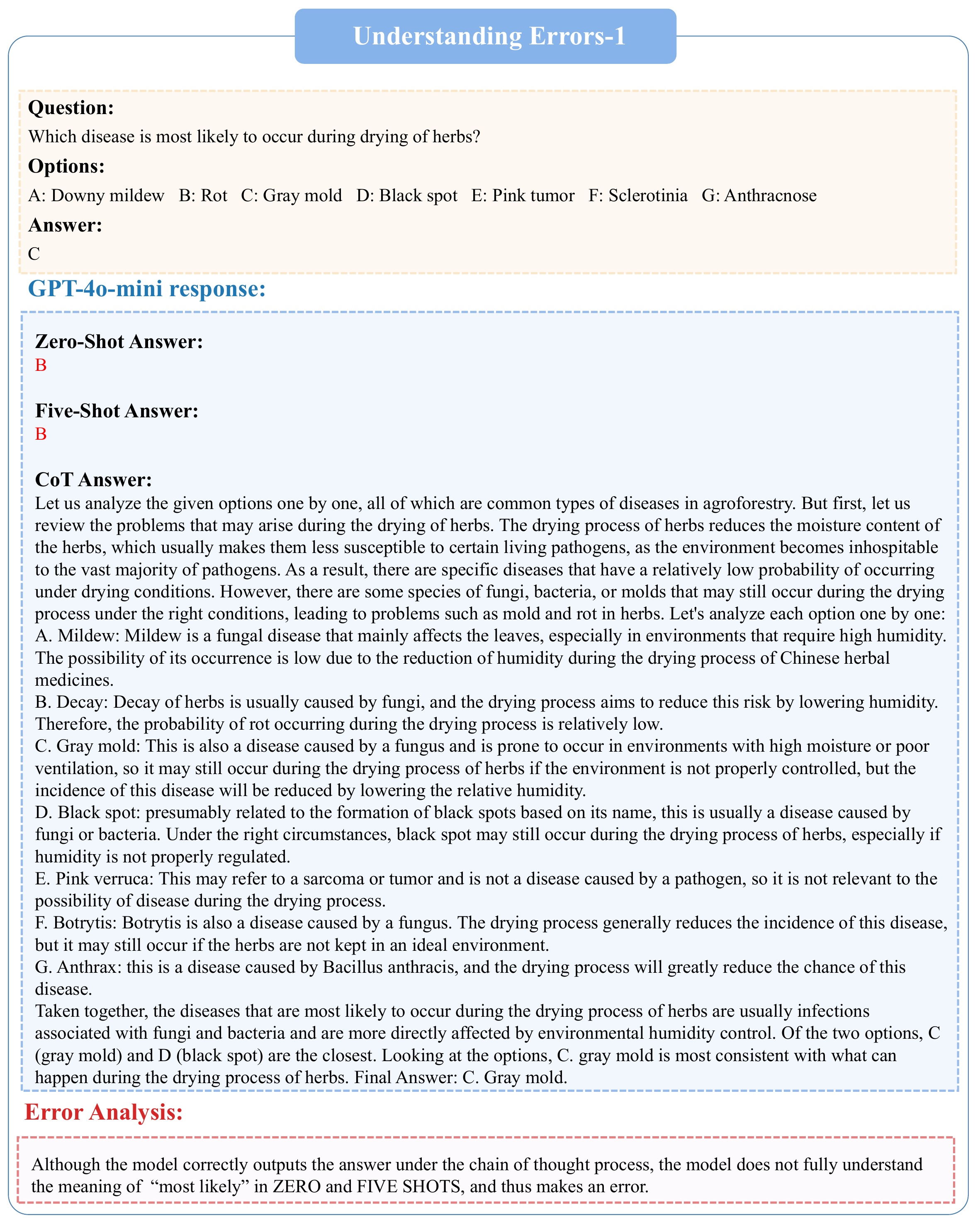}
    \caption{A sample for understanding errors.}
    \label{fig:understanding-error-1}
\end{figure}

\begin{figure}[htbp]
    \centering
   \includegraphics[width=\linewidth]{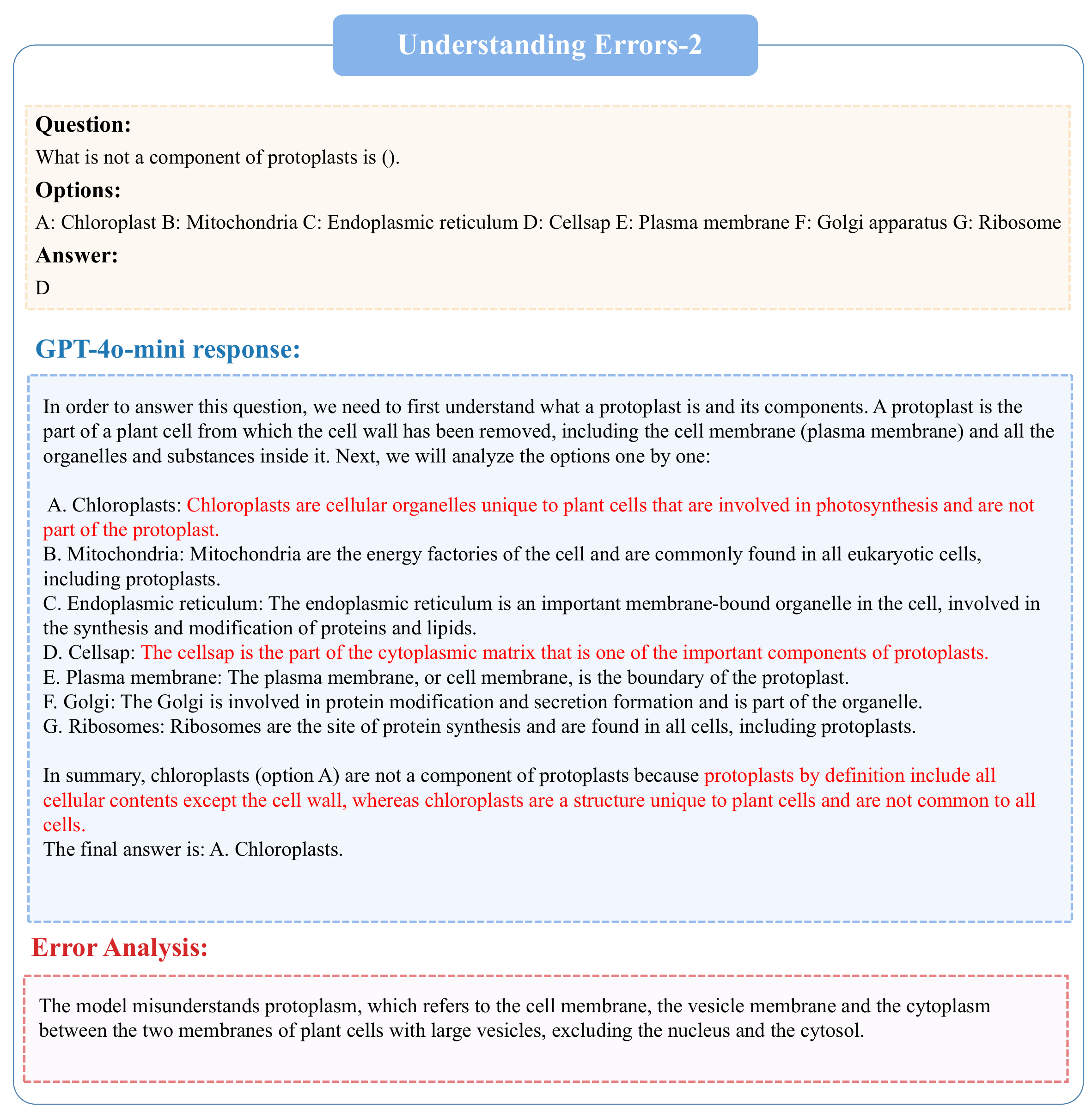}
    \caption{A sample for understanding errors.}
    \label{fig:understanding-error-2}
\end{figure}

\begin{figure}[htbp]
    \centering
   \includegraphics[width=\linewidth]{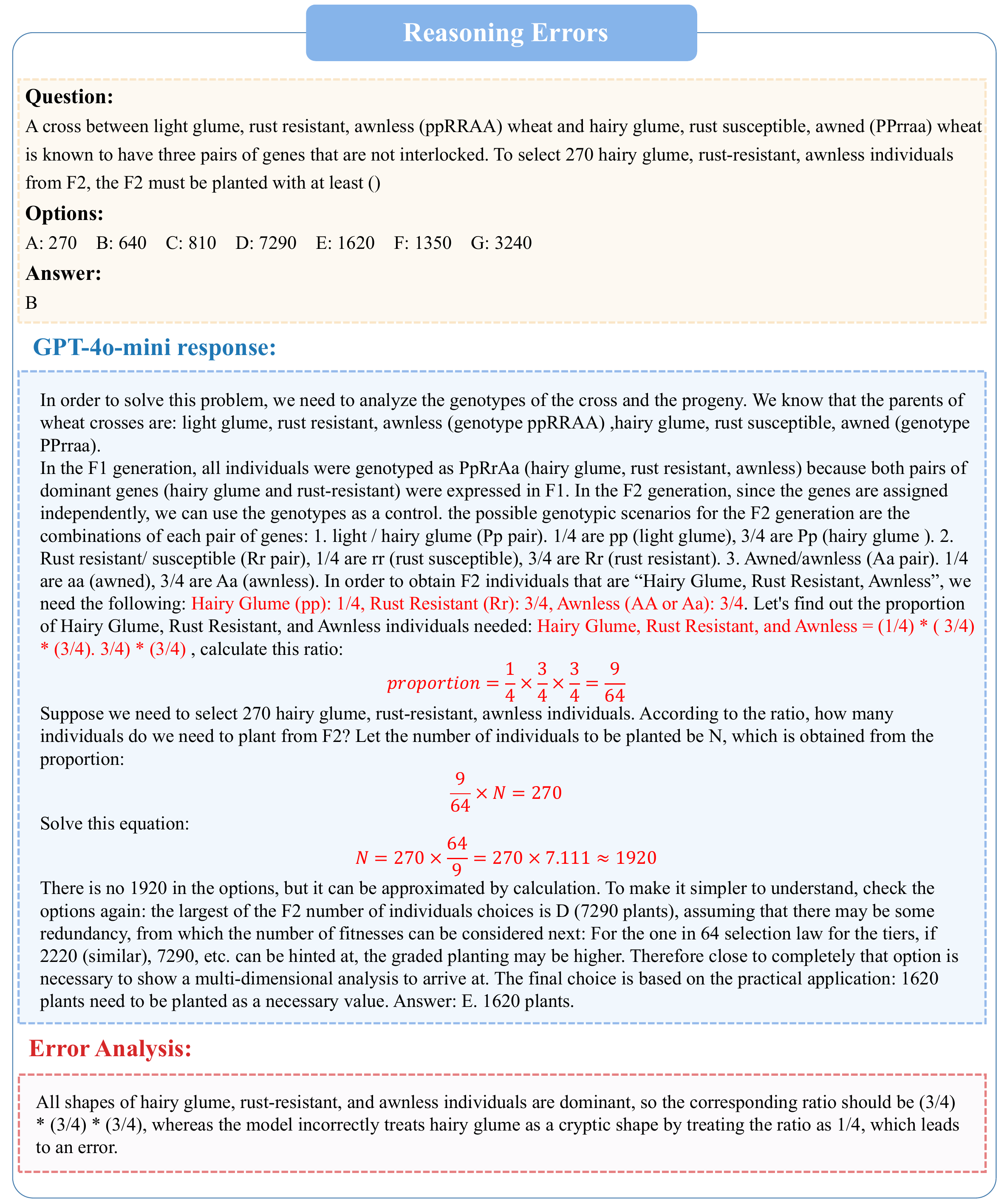}
    \caption{A sample for reasoning errors.}
    \label{fig:reasoning-error}
\end{figure}

Representative examples of knowledge errors, understanding errors, and reasoning errors are shown in Figure~\ref{fig:knowledge-error}, Figure~\ref{fig:understanding-error-1},  Figure~\ref{fig:understanding-error-2}, and Figure~\ref{fig:reasoning-error}, respectively.

\section{Limitations}\label{app-limitation}
Although AgriEval is the most extensive Chinese agricultural benchmark for LLMs, covers a wide range of agricultural domains and comprehensively evaluates various cognitive abilities of LLMs, it has several limitations: (1) AgriEval is collected from Chinese university- and graduate-level exam questions, which restricts its applicability to multilingual tasks. (2) It contains a few questions about drones and agricultural machinery, which are crucial for tool usage in smart agriculture, an essential real-world application. (3) In context-based generation tasks, AgriEval primarily assesses the ability to generate professional agricultural responses based on existing information rather than exploring the model's creative potential, which could contribute to a new variety of discoveries. We will continue developing a more comprehensive and advanced benchmark.

Another notable limitation lies in the evaluation metric. For generation-level tasks, we adopt Rouge-L as the primary metric. However, Rouge-L relies on character-level overlap, which may fail to fully capture LLMs' true performance in agriculture, as models often produce semantically correct but lexically diverse responses.

\begin{table}[]
\centering
\caption{Zero-shot performance on cognitive-specific multi-choice tasks in AgriEval.}
\label{tab:appendix_domain_choice_main_results_zero-shot_choice}
\resizebox{\textwidth}{!}{%
\begin{tabular}{cccccccccccccc}
\toprule
\multirow{2}{*}{\textbf{Model}} & \multicolumn{4}{c}{\textbf{Memorization}}                 & \multicolumn{3}{c}{\textbf{Understanding}} & \multicolumn{5}{c}{\textbf{Inference}}                                   & \multirow{2}{*}{\textbf{Overall}} \\ \cmidrule(lr){2-5} \cmidrule(lr){6-8} \cmidrule(lr){9-13}
                                & \textbf{M-P} & \textbf{M-R} & \textbf{M-E} & \textbf{M-T} & \textbf{U-I} & \textbf{U-S} & \textbf{U-V} & \textbf{I-D} & \textbf{I-N} & \textbf{I-P} & \textbf{I-S} & \textbf{I-G} &                                   \\ \midrule
DeepSeek-LLM-7B-Base            & 17.01        & 14.75        & 22.35        & 19.70         & 15.77        & 19.13        & 38.57        & 16.38        & 12.16        & 22.86        & 14.53        & 12.15        & 20.03                             \\
Llama2-13B-Chat                 & 17.69        & 15.30         & 18.56        & 12.12        & 15.55        & 15.48        & 46.61        & 19.77        & 14.88        & 17.34        & 18.07        & 13.40         & 21.15                             \\
Baichuan2-7B-Base               & 19.77        & 18.03        & 21.89        & 21.97        & 18.48        & 20.72        & 47.59        & 21.26        & 15.67        & 25.19        & 20.88        & 14.64        & 23.84                             \\
Llama2-7B-Chat                  & 21.06        & 21.31        & 19.55        & 30.30         & 18.79        & 20.77        & 46.83        & 25.31        & 17.61        & 18.4         & 14.65        & 19.63        & 23.85                             \\
Baichuan2-13B-Base              & 22.79        & 22.40         & 28.41        & 31.06        & 20.55        & 24.50         & 44.96        & 25.14        & 18.82        & 30.15        & 24.79        & 21.18        & 26.09                             \\
\bottomrule
\end{tabular}
}
\end{table}

\begin{table}[]
\centering
\caption{Zero-shot performance on domain-specific multiple-choice tasks in AgriEval.}
\label{tab:appendix_domain_zero-shot_choice_main_results}
\resizebox{0.85\textwidth}{!}{%
\begin{tabular}{cccccccc}
\toprule
\textbf{Model}            & \textbf{PP} & \textbf{Aqua} & \textbf{GS} & \textbf{TCH}   & \textbf{Fore} & \textbf{AST}   & \textbf{Overall} \\ \midrule
DeepSeek-LLM-7B-Base      & 20.11            & 20.96       & 15.13         & 16.97 & 20.96    & 20.04 & 20.03   \\
Llama2-13B-Chat           & 21.36            & 19.87       & 14.87         & 17.23 & 20.50    & 21.77 & 21.15   \\
Baichuan2-7B-Base         & 24.07            & 23.91       & 13.33         & 21.79 & 22.40    & 23.99 & 23.84   \\
Llama2-7B-Chat            & 24.12            & 22.95       & 15.64         & 17.92 & 21.74    & 25.03 & 23.85   \\
Baichuan2-13B-Base        & 26.07            & 26.54       & 19.23         & 24.12 & 28.81    & 26.20 & 26.09   \\
Llama2-70B-Chat           & 27.35            & 25.00       & 26.15         & 21.71 & 24.75    & 27.63 & 27.05   \\
Mistral-7B-Instruct       & 29.40            & 32.44       & 33.33         & 21.53 & 30.45    & 26.81 & 29.10   \\
\bottomrule
\end{tabular}
}
\end{table}

\begin{table}[]
\centering
\caption{Few-shot performance on cognitive-specific multi-choice tasks in AgriEval. {\increase{}}/{\decrease{}} represents the performance increase/decrease compared to the zero-shot setting.}
\label{tab:appendix_domain_choice_main_results_few-shot_choice}
\resizebox{\textwidth}{!}{%

\begin{tabular}{cccccccccccccc}
\toprule
\multicolumn{1}{c}{\multirow{2}{*}{\textbf{Model}}} & \multicolumn{4}{c}{\textbf{Memorization}}                 & \multicolumn{3}{c}{\textbf{Understanding}} & \multicolumn{5}{c}{\textbf{Inference}}                                                       & \multirow{2}{*}{\textbf{Overall}} \\ \cmidrule(lr){2-5} \cmidrule(lr){6-8} \cmidrule(lr){9-13}
\multicolumn{1}{c}{}                                & \textbf{M-T} & \textbf{M-P} & \textbf{M-R} & \textbf{M-E} & \textbf{U-V} & \textbf{U-I} & \textbf{U-S} & \textbf{I-P} & \textbf{I-N} & \textbf{I-D} & \textbf{I-S} & \multicolumn{1}{c}{\textbf{I-G}} &                                   \\ \midrule 
Llama2-7B-Chat                                      & 8.39         & 13.11        & 8.18         & 13.64        & 7.31         & 8.16         & 26.86        & 11.41        & 11.64        & 9.06         & 15.26        & 5.92                             & 11.02\decrease{}                                    \\
DeepSeek-LLM-7B-Base                                & 16.14        & 12.57        & 16.36        & 11.36        & 16.05        & 16.34        & 41.63        & 15.14        & 14.05        & 19.18        & 12.33        & 14.02                            & 19.75\decrease{}                                    \\
Llama2-13B-Chat                                     & 15.36        & 14.21        & 13.41        & 12.88        & 16.01        & 16.01        & 49.84        & 15.72        & 13.89        & 16.42        & 14.77        & 13.40                            & 19.86\decrease{}                                    \\
Baichuan2-7B-Base                                   & 17.16        & 12.02        & 17.92        & 16.67        & 15.40        & 17.40        & 46.97        & 16.96        & 14.62        & 16.70        & 13.43        & 16.82                            & 20.52\decrease{}                                    \\
\bottomrule
\end{tabular}

}
\end{table}

\begin{table}[]
\centering
\caption{Few-shot performance on domain-specific multiple-choice tasks in AgriEval. {\increase{}}/{\decrease{}} represents the performance increase/decrease compared to the zero-shot setting.}
\label{tab:appendix_domain_few-shot_choice_main_results}
\resizebox{0.85\textwidth}{!}{%
\begin{tabular}{cccccccc}
\toprule
\textbf{Model}            & \textbf{PP} & \textbf{Aqua} & \textbf{GS} & \textbf{TCH}   & \textbf{Fore} & \textbf{AST}   & \textbf{Overall} \\ \midrule
Llama2-7B-Chat            & 11.45            & 7.49        & 10.90         & 8.21  & 7.40     & 10.16 & 11.02\decrease{}   \\
Llama2-13B-Chat           & 20.17            & 16.37       & 17.63         & 13.85 & 15.78    & 20.02 & 19.75\decrease{}   \\
DeepSeek-LLM-7B-Base      & 20.12            & 15.85       & 18.33         & 14.36 & 18.40    & 20.38 & 19.86\decrease{}   \\
Baichuan2-7B-Base         & 20.83            & 16.71       & 21.09         & 13.33 & 16.96    & 20.71 & 20.52\decrease{}   \\
Baichuan2-13B-Chat        & 21.74            & 14.81       & 18.72         & 15.64 & 15.32    & 20.43 & 21.02\decrease{}   \\
Llama3-8B-Instruct        & 21.83            & 17.83       & 20.06         & 14.36 & 17.35    & 22.27 & 21.48\decrease{}   \\
Baichuan2-13B-Base        & 21.74            & 17.57       & 20.19         & 17.18 & 20.63    & 22.01 & 21.52\decrease{}   \\
\bottomrule
\end{tabular}
}
\end{table}

\begin{table}[]
\caption{CoT performance on cognitive-specific multi-choice tasks in AgriEval. {\increase{}}/{\decrease{}} represents the performance increase/decrease compared to the zero-shot setting.}
\label{tab:appendix_domain_choice_main_results_cot_choice}
\resizebox{\textwidth}{!}{%
\begin{tabular}{cccccccccccccc}
\toprule
\multirow{2}{*}{\textbf{Model}} & \multicolumn{4}{c}{\textbf{Memorization}}                 & \multicolumn{3}{c}{\textbf{Understanding}} & \multicolumn{5}{c}{\textbf{Inference}}                                   & \multirow{2}{*}{\textbf{Overall}} \\ \cmidrule(lr){2-5} \cmidrule(lr){6-8} \cmidrule(lr){9-13}
                                & \textbf{M-P} & \textbf{M-R} & \textbf{M-E} & \textbf{M-T} & \textbf{U-I} & \textbf{U-S} & \textbf{U-V} & \textbf{I-D} & \textbf{I-N} & \textbf{I-P} & \textbf{I-S} & \textbf{I-G} &                                   \\ \midrule
Llama2-13B-Chat                                     & 15.07        & 16.94        & 13.86        & 24.24        & 15.68        & 16.72        & 47.54        & 14.06        & 14.83        & 14.93        & 12.70        & 17.45                            & 19.52\decrease{}                \\
Llama2-70B-Chat                                     & 14.88        & 19.67        & 14.66        & 13.64        & 15.80        & 15.11        & 48.34        & 15.14        & 13.21        & 20.17        & 19.05        & 12.15                            & 19.82\decrease{}                \\
Llama3-8B-Instruct                                  & 15.42        & 16.94        & 15.72        & 12.88        & 15.39        & 17.04        & 48.50        & 15.38        & 14.05        & 17.69        & 14.53        & 12.77                            & 20.05\decrease{}                \\
Marco-o1                                            & 15.28        & 15.85        & 14.47        & 14.39        & 15.14        & 15.01        & 51.23        & 15.14        & 16.14        & 16.77        & 14.90        & 15.58                            & 20.26\decrease{}                \\
\bottomrule
\end{tabular}
}
\end{table}

\begin{table}[]
\centering
\caption{CoT performance on domain-specific multiple-choice tasks in AgriEval. {\increase{}}/{\decrease{}} represents the performance increase/decrease compared to the zero-shot setting.}
\label{tab:appendix_domain_cot_choice_main_results}
\resizebox{0.85\textwidth}{!}{%
\begin{tabular}{cccccccc}
\toprule
\textbf{Model}            & \textbf{PP} & \textbf{Aqua} & \textbf{GS} & \textbf{TCH}   & \textbf{Fore} & \textbf{AST}   & \textbf{Overall} \\ \midrule
Llama2-13B-Chat           & 19.97            & 16.19       & 17.44         & 16.15 & 15.26    & 19.43 & 19.52\decrease{}   \\
Llama2-70B-Chat           & 20.16            & 15.76       & 19.62         & 13.85 & 15.32    & 20.36 & 19.82\decrease{}   \\
Llama3-8B-Instruct        & 20.58            & 16.62       & 16.79         & 13.59 & 16.50    & 19.82 & 20.05\decrease{}   \\
Marco-o1                  & 20.71            & 17.31       & 18.59         & 14.10 & 16.83    & 19.91 & 20.26\decrease{}   \\
DeepSeek-LLM-7B-Base      & 20.27            & 18.35       & 22.24         & 16.67 & 22.46    & 21.70 & 20.49\increase{}    \\
Llama3-70B-Instruct       & 21.08            & 16.45       & 20.71         & 13.59 & 16.24    & 19.48 & 20.54\decrease{}   \\
Llama2-7B-Chat            & 22.33            & 15.85       & 20.77         & 15.38 & 21.55    & 23.00 & 22.08\decrease{}   \\
\bottomrule
\end{tabular}
}
\end{table}

\begin{table}[]
\centering
\caption{Zero-shot performance on cognitive levels in AgriEval generation tasks.}
\label{tab:appendix_cognitive_domain_main_results_zero-shot_generation}
\resizebox{0.65\textwidth}{!}{%
\begin{tabular}{ccccc}
\toprule
\textbf{Model}            & \textbf{G-QA} & \textbf{G-PS} & \textbf{G-CA} & \textbf{Overall} \\ \midrule
Marco-o1                  & 0.0069        & 0.0187        & 0.0041        & 0.0065           \\
Llama2-13B-Chat           & 0.0257        & 0.0373        & 0.0116        & 0.0164           \\
Llama2-70B-Chat           & 0.0264        & 0.0477        & 0.0156        & 0.0212           \\
Llama2-7B-Chat            & 0.0355        & 0.0564        & 0.0218        & 0.0278           \\
Llama3-8B-Instruct        & 0.0521        & 0.0936        & 0.0256        & 0.0376           \\
Llama3-70B-Instruct       & 0.0525        & 0.0915        & 0.0301        & 0.0409           \\
Mistral-7B-Instruct       & 0.0968        & 0.1485        & 0.0450        & 0.0639           \\
\bottomrule
\end{tabular}
}
\end{table}

\begin{table}[]
\centering
\caption{Comparison of CoT and zero-shot performance across multiple-choice formats, including single-answer (SA), multi-answer (MA), and true/false (T/F) questions. {\increase{}}/{\decrease{}} represents the performance increase/decrease compared to the zero-shot setting.}
\label{tab:appendix_choice_formats}
\begin{minipage}{0.48\linewidth}
\resizebox{\linewidth}{!}{%
\begin{tabular}{cccccc}
\toprule
\textbf{Model}                                      & \textbf{CoT} & \textbf{SA} & \textbf{MA} & \textbf{T/F} & \textbf{Overall} \\ \midrule
\multirow{2}{*}{DeepSeek-LLM-7B-Base}      & \xmark                   & 17.59                             & 15.31                            & 38.63                          & 20.03                       \\
                                           & \cmark                   & 19.40\increase{}                             & 14.21\decrease{}                             & 31.73\decrease{}                           & 20.49\increase{}                        \\ \midrule
\multirow{2}{*}{Llama2-13B-Chat}           & \xmark                   & 18.53                             & 8.37                             & 46.72                          & 21.15                       \\
                                           & \cmark                   & 14.49\decrease{}                              & 19.28\increase{}                             & 47.66\increase{}                           & 19.52\decrease{}                        \\ \midrule
\multirow{2}{*}{Baichuan2-7B-Base}         & \xmark                   & 20.02                             & 18.60                            & 47.69                          & 23.84                       \\
                                           & \cmark                   & 23.04\increase{}                              & 17.05\decrease{}                             & 45.07\decrease{}                           & 25.42\increase{}                        \\ \midrule
\multirow{2}{*}{Llama2-7B-Chat}            & \xmark                   & 22.89                             & 5.54                             & 46.94                          & 23.85                       \\
                                           & \cmark                   & 19.70\decrease{}                              & 10.15\increase{}                             & 47.50\increase{}                           & 22.08\decrease{}                        \\ \midrule
\multirow{2}{*}{Baichuan2-13B-Base}        & \xmark                   & 24.26                             & 17.12                            & 45.02                          & 26.09                       \\
                                           & \cmark                   & 26.08\increase{}                              & 17.00\decrease{}                             & 39.01\decrease{}                           & 26.64\increase{}                        \\ \midrule
\multirow{2}{*}{Llama2-70B-Chat}           & \xmark                   & 27.03                             & 1.62                             & 52.02                          & 27.05                       \\
                                           & \cmark                   & 14.93\decrease{}                              & 17.05\increase{}                             & 48.44\decrease{}                           & 19.82\decrease{}                        \\ \midrule
\multirow{2}{*}{Mistral-7B-Instruct}       & \xmark                   & 29.07                             & 8.91                             & 48.97                          & 29.10                       \\
                                           & \cmark                   & 23.11\decrease{}                              & 14.14\increase{}                            & 51.49\increase{}                           & 25.55\decrease{}                        \\ \midrule
\multirow{2}{*}{Llama3-8B-Instruct}        & \xmark                   & 32.13                             & 10.74                            & 47.45                          & 31.38                       \\
                                           & \cmark                   & 14.95\decrease{}                              & 19.04\increase{}                             & 48.58\increase{}                           & 20.05\decrease{}                        \\ \midrule
\multirow{2}{*}{Qwen1.5-1.8B-Chat}         & \xmark                   & 32.36                             & 19.07                            & 44.29                          & 32.25                       \\
                                           & \cmark                   & 27.57\decrease{}                              & 22.52\increase{}                             & 45.14\increase{}                           & 29.27\decrease{}                        \\ \midrule
\multirow{2}{*}{Baichuan2-7B-Chat}         & \xmark                   & 33.55                             & 8.40                             & 50.14                          & 32.46                       \\
                                           & \cmark                   & 29.20\decrease{}                              & 23.66\increase{}                             & 48.41\decrease{}                           & 31.07\decrease{}                        \\ \midrule
\multirow{2}{*}{ChatGLM2-6B}               & \xmark                   & 33.54                             & 13.44                            & 47.74                          & 32.80                       \\
                                           & \cmark                   & 30.57\decrease{}                              & 21.46\increase{}                             & 46.16\decrease{}                           & 31.56\decrease{}                        \\ \midrule
\multirow{2}{*}{InternLM2-Chat-7B}         & \xmark                   & 34.53                             & 12.27                            & 49.40                          & 33.58                       \\
                                           & \cmark                   & 32.17\decrease{}                              & 26.08\increase{}                             & 48.20\decrease{}                           & 33.49\decrease{}                        \\ \midrule
\multirow{2}{*}{DeepSeek-LLM-7B-Chat}      & \xmark                   & 35.10                             & 8.14                             & 51.52                          & 33.76                       \\
                                           & \cmark                   & 33.08\decrease{}                              & 22.44\increase{}                             & 51.41\decrease{}                           & 34.22\increase{}                        \\ \midrule
\multirow{2}{*}{Qwen1.5-4B-Chat}           & \xmark                   & 34.93                             & 21.13                            & 43.18                          & 34.20                       \\
                                           & \cmark                   & 33.55\decrease{}                              & 28.47\increase{}                            & 37.28\decrease{}                           & 33.41\decrease{}                        \\ \midrule
\multirow{2}{*}{Phi-3.5-Mini-Instruct}     & \xmark                   & 36.96                             & 6.75                             & 46.21                          & 34.22                       \\
                                           & \cmark                   & 29.23\decrease{}                              & 27.60\increase{}                            & 47.33\increase{}                          & 31.4\decrease{}                         \\ \midrule
\multirow{2}{*}{Mistral-Nemo-Instruct}     & \xmark                   & 36.04                             & 9.40                             & 50.03                          & 34.39                       \\
                                           & \cmark                   & 23.70\decrease{}                             & 15.62\increase{}                            & 50.11\increase{}                          & 26.19\decrease{}                       \\ \midrule
\multirow{2}{*}{Qwen-7B-Chat}              & \xmark                   & 35.75                             & 17.78                            & 47.91                          & 35.03                       \\
                                           & \cmark                   & 31.49\decrease{}                              & 27.23\increase{}                            & 46.47\decrease{}                           & 33.01\decrease{}                        \\ \midrule
\multirow{2}{*}{Baichuan2-13B-Chat}        & \xmark                   & 37.99                             & 9.71                             & 47.44                          & 35.53                       \\
                                           & \cmark                   & 34.73\decrease{}                             & 25.16\increase{}                            & 47.50\increase{}                          & 35.22\decrease{}                       \\ \midrule
\multirow{2}{*}{ChatGLM3-6b}               & \xmark                   & 36.97                             & 13.48                            & 49.82                          & 35.55                       \\
                                           & \cmark                   & 27.60\decrease{}                             & 22.02\increase{}                            & 46.82\decrease{}                          & 29.45\decrease{}                       \\ \midrule
\multirow{2}{*}{Qwen1.5-7B-Chat}           & \xmark                   & 35.61                             & 28.94                            & 48.68                          & 36.49                       \\
                                           & \cmark                   & 36.69\increase{}                             & 37.24\increase{}                            & 48.73\increase{}                          & 38.39\increase{}                       \\ \midrule
\multirow{2}{*}{Phi-3-Medium-4k-Instruct}  & \xmark                   & 39.06                             & 11.31                            & 49.96                          & 36.87                       \\
                                           & \cmark                   & 33.46\decrease{}                             & 31.15\increase{}                            & 49.77\decrease{}                          & 35.38\decrease{}                       \\ \midrule
\multirow{2}{*}{Marco-o1}                  & \xmark                   & 37.55                             & 23.34                            & 49.82                          & 37.32                       \\
                                           & \cmark                   & 14.41\decrease{}                             & 19.93\decrease{}                            & 51.35\increase{}                         & 20.26\decrease{}                       \\    
\bottomrule
\end{tabular}
}
\end{minipage}
\hfill
\begin{minipage}{0.49\linewidth}
\resizebox{\linewidth}{!}{%
\begin{tabular}{cccccc}
\toprule
\textbf{Model}                                      & \textbf{CoT} & \textbf{SA} & \textbf{MA} & \textbf{T/F} & \textbf{Overall} \\ \midrule

\multirow{2}{*}{Qwen-14B-Chat}             & \xmark                   & 38.04                             & 26.38                            & 51.05                          & 38.25                       \\
                                           & \cmark                   & 40.01\increase{}                             & 34.43\increase{}                            & 49.36\decrease{}                          & 40.55\increase{}                       \\ \midrule
\multirow{2}{*}{Qwen2.5-Coder-7B-Instruct} & \xmark                   & 39.16                             & 26.19                            & 48.39                          & 38.69                       \\
                                           & \cmark                   & 37.77\decrease{}                             & 36.89\increase{}                            & 50.86\increase{}                          & 39.44\increase{}                       \\ \midrule
\multirow{2}{*}{Qwen2.5-3B-Instruct}       & \xmark                   & 42.38                             & 13.25                            & 50.79                          & 39.67                       \\
                                           & \cmark                   & 37.83\decrease{}                             & 31.14\increase{}                            & 52.31\increase{}                          & 38.9\decrease{}                        \\ \midrule
\multirow{2}{*}{Yi-1.5-9B-Chat}            & \xmark                   & 40.25                             & 28.54                            & 58.83                          & 41.20                       \\
                                           & \cmark                   & 40.21\decrease{}                             & 34.33\increase{}                            & 54.22\decrease{}                          & 41.29\increase{}                       \\ \midrule
\multirow{2}{*}{InternLM2.5-20b-Chat}      & \xmark                   & 42.47                             & 29.06                            & 53.96                          & 42.26                       \\
                                           & \cmark                   & 39.74\decrease{}                             & 34.82\increase{}                            & 55.00\increase{}                          & 41.18\decrease{}                       \\ \midrule
\multirow{2}{*}{Llama3-70B-Instruct}       & \xmark                   & 44.52                             & 29.10                            & 49.46                          & 43.16                       \\
                                           & \cmark                   & 14.68\decrease{}                             & 19.39\decrease{}                            & 51.00\increase{}                          & 20.54\decrease{}                       \\ \midrule
\multirow{2}{*}{GLM-4-9b-Chat}             & \xmark                   & 44.38                             & 31.97                            & 51.61                          & 43.72                       \\
                                           & \cmark                   & 43.15\decrease{}                             & 37.52\increase{}                            & 52.99\increase{}                          & 43.78\increase{}                       \\ \midrule
\multirow{2}{*}{InternLM2.5-7B-Chat}       & \xmark                   & 46.15                             & 20.99                            & 52.87                          & 43.75                       \\
                                           & \cmark                   & 39.09\decrease{}                             & 33.04\increase{}                            & 52.43\decrease{}                          & 40.04\decrease{}              \\ \midrule
\multirow{2}{*}{Qwen2-72B-Instruct}        & \xmark                   & 62.06                             & 60.11                            & 68.81                          & 62.72              \\
                                           & \cmark                   & 50.45\decrease{}                             & 52.32\decrease{}                            & 65.88\decrease{}                          & 52.72\decrease{}                       \\ \midrule
\multirow{2}{*}{Qwen1.5-14B-Chat}          & \xmark                   & 46.19                             & 33.75                            & 50.37                          & 45.11                       \\
                                           & \cmark                   & 42.82\decrease{}                             & 41.00\increase{}                            & 43.04\decrease{}                          & 42.59\decrease{}                       \\ \midrule
\multirow{2}{*}{Qwen1.5-32B-Chat}          & \xmark                   & 45.74                             & 41.28                            & 54.47                          & 46.33                       \\
                                           & \cmark                   & 43.45\decrease{}                             & 46.23\increase{}                            & 47.13\decrease{}                          & 44.30\decrease{}                        \\ \midrule
\multirow{2}{*}{Qwen-72B-Chat}             & \xmark                   & 45.30                             & 47.35                            & 56.19                          & 47.05                       \\
                                           & \cmark                   & 49.23\increase{}                             & 37.41\decrease{}                            & 55.03\decrease{}                          & 48.42\increase{}                       \\ \midrule
\multirow{2}{*}{Qwen2.5-7B-Instruct}       & \xmark                   & 49.46                             & 35.58                            & 53.79                          & 48.21                       \\
                                           & \cmark                   & 42.39\decrease{}                             & 42.10\increase{}                            & 53.38\decrease{}                          & 43.82\decrease{}                       \\ \midrule
\multirow{2}{*}{Qwen2-7B-Instruct}         & \xmark                   & 48.75                             & 41.51                            & 56.33                          & 48.83                       \\
                                           & \cmark                   & 38.63\decrease{}                             & 24.83\decrease{}                            & 56.29\decrease{}                          & 39.33\decrease{}                       \\ \midrule
\multirow{2}{*}{Qwen2.5-14B-Instruct}      & \xmark                   & 50.49                             & 37.73                            & 55.82                          & 49.53                       \\
                                           & \cmark                   & 47.06\decrease{}                             & 45.82\increase{}                            & 57.93\increase{}                          & 48.39\decrease{}                       \\ \midrule
\multirow{2}{*}{Yi-1.5-34b-Chat}           & \xmark                   & 51.67                             & 41.82                            & 62.44                          & 51.83                       \\
                                           & \cmark                   & 43.72\decrease{}                             & 42.90\increase{}                            & 59.75\decrease{}                          & 45.77\decrease{}                       \\ \midrule
\multirow{2}{*}{Qwen1.5-72B-Chat}          & \xmark                   & 52.63                             & 51.74                            & 56.06                          & 52.98                       \\
                                           & \cmark                   & 46.46\decrease{}                             & 45.90\decrease{}                            & 56.55\increase{}                          & 47.81\decrease{}                       \\ \midrule
\multirow{2}{*}{Qwen2.5-32B-Instruct}      & \xmark                   & 57.02                             & 47.26                            & 61.61                          & 56.35                       \\
                                           & \cmark                   & 48.07\decrease{}                             & 48.69\increase{}                            & 61.28\decrease{}                          & 49.99\decrease{}                       \\ \midrule
\multirow{2}{*}{DeepSeek-V3}               & \xmark                   & 60.16                             & 39.85                            & 59.79                          & 57.43                       \\
                                           & \cmark                   & 54.97\decrease{}                             & 42.83\increase{}                            & 62.29\increase{}                          & 54.23\decrease{}                       \\ \midrule
\multirow{2}{*}{Qwen2.5-72B-Instruct}      & \xmark                   & 60.81                             & 54.99                            & 62.80                          & 60.32                       \\
                                           & \cmark                   & 47.78\decrease{}                             & 42.12\decrease{}                            & 62.05\decrease{}                          & 49.05\decrease{}                       \\ \midrule
\multirow{2}{*}{GPT-3.5-Turbo}             & \xmark                   & 36.26                             & 9.15                             & 49.21                          & 34.43                       \\
                                           & \cmark                   & 34.64\decrease{}                             & 35.41\increase{}                            & 49.92\increase{}                          & 36.80\increase{}                       \\ \midrule
\multirow{2}{*}{GPT-4o-mini}               & \xmark                   & 50.38                             & 36.09                            & 48.13                          & 48.19                       \\
                                           & \cmark                   & 41.63\decrease{}                             & 41.63\increase{}                            & 54.52\increase{}                          & 43.37\decrease{}             \\         
\bottomrule
\end{tabular}
}
\end{minipage}

\label{aptable5}
\end{table}

\end{document}